%% file: main.tex
\documentclass[runningheads]{llncs}

\usepackage[review,year=2026,ID=12]{eccv}
\usepackage{lipsum}
\usepackage{booktabs}
\usepackage{multirow}
\usepackage{array}
\usepackage{graphicx}
\usepackage{colortbl}
\usepackage{xcolor}
\usepackage{wrapfig}
\usepackage{hyperref}
\definecolor{cvprblue}{rgb}{0.21,0.49,0.74}
\definecolor{lightpurblue}{RGB}{243, 249, 255}
\definecolor{midpurblue}{RGB}{205, 227, 255}
\usepackage{pifont}

\usepackage{eccvabbrv}

\usepackage{graphicx}
\usepackage{booktabs}

\usepackage[accsupp]{axessibility}  % Improves PDF readability for those with disabilities.

\usepackage{hyperref}

\usepackage{orcidlink}
\usepackage[capitalize]{cleveref}

\begin{document}

%\title{Repurposing Geometric Foundation Models as Multi-view Diffusion Models} 
\title{Repurposing Geometric Foundation Models for Multi-view Diffusion} 

\titlerunning{Geometric Latent Diffusion}

\author{
Wooseok Jang$^{1}$ \and
Seonghu Jeon$^{1}$ \and
Jisang Han$^{1}$ \and
Jinhyeok Choi$^{1}$ \and \\
Minkyung Kwon$^{1}$ \and
Seungryong Kim$^{1}$ \and
Saining Xie$^{2}$ \and
Sainan Liu$^{3}$
}
%\vspace{-40pt}

\authorrunning{W. Jang et al.}

\institute{$^{1}$KAIST AI \quad $^{2}$New York University \quad $^{3}$Intel Labs \\
\textcolor{Blue}{\href{https://cvlab-kaist.github.io/GLD}{https://cvlab-kaist.github.io/GLD}}}
\maketitle

\begin{figure}
\vspace{-25pt}
    \includegraphics[width=\linewidth]{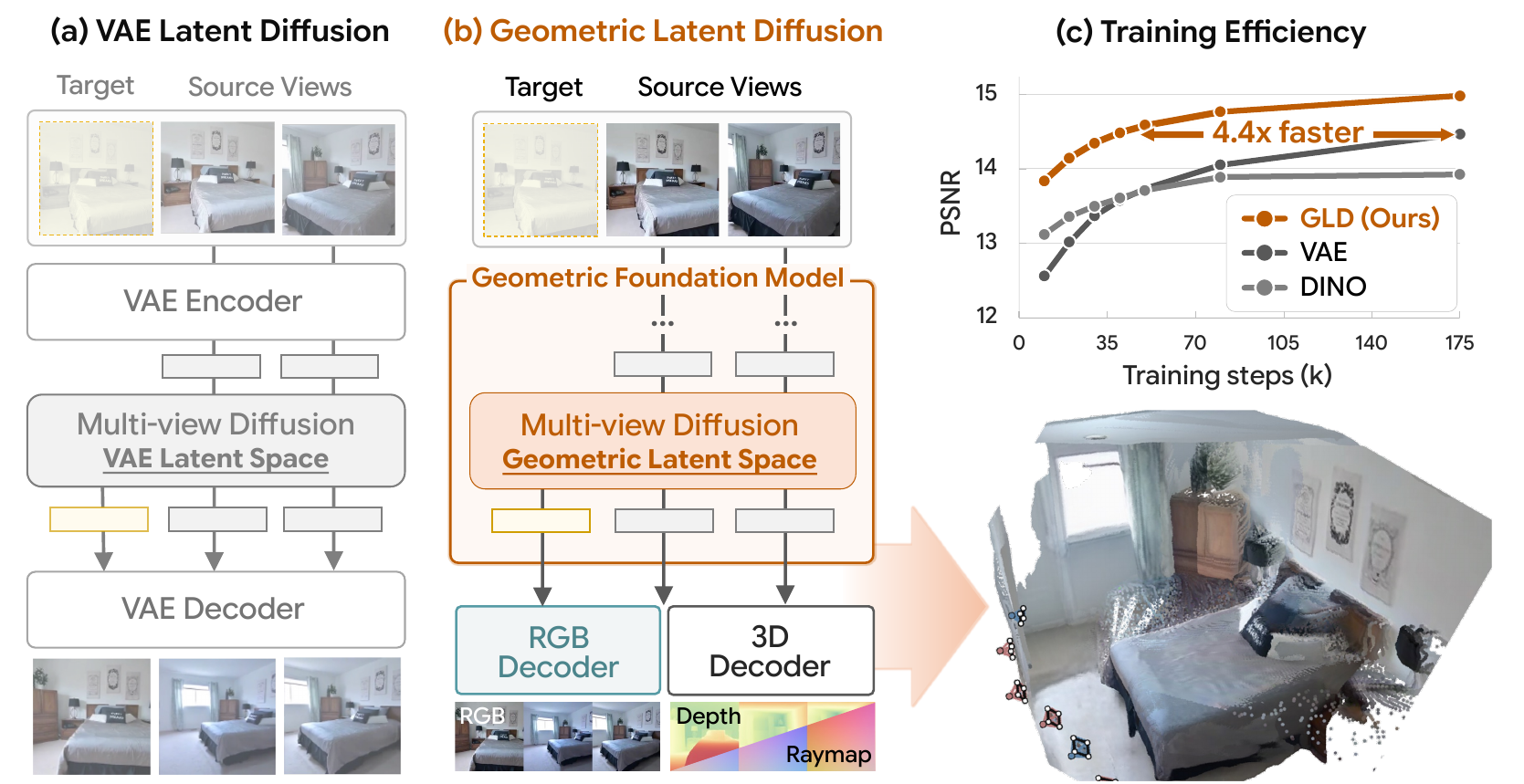}
    \vspace{-15pt}
    \captionof{figure}{\textbf{(a)} Prior multi-view diffusion operates in a view-independent VAE~\cite{rombach2022high} latent space.
\textbf{(b)} \textbf{Geometric Latent Diffusion (GLD)} instead diffuses in a geometrically consistent feature space of geometric foundation models~\cite{da3,wang2025vggt}, enabling both RGB and geometry decoding.
\textbf{(c)} GLD converges $\mathbf{4.4\times}$ faster than VAE~\cite{rombach2022high} and DINO~\cite{rae}.}
    \label{fig:teaser}
    \vspace{-30pt}
\end{figure}

\input{sec/0_abstract}
\input{sec/1_introduction}
\input{sec/2_relwork}

\input{sec/3_preliminary}
% \input{sec/4_analysis}
\input{sec/5_method}

\input{sec/6_experiment}
\input{sec/7_conclusion}

% \clearpage  % TODO FINAL: This \clearpage needs to be removed from both review and camera-ready versions.

% \section*{Acknowledgements}
% Please insert your acknowledgments here.

\clearpage
\bibliographystyle{splncs04}
\bibliography{main}

\clearpage
\appendix

%\begin{center}
%    {\Large\bfseries Repurposing Geometric Foundation Models for Multi-view Diffusion\par}
%    \vspace{0.4em}
%    {\Large\ttfamily -- Supplementary Materials --\par}
%\end{center}

% \vspace{1.2em}

\input{sec/appendix}
\end{document}

%% file: sec/0_abstract.tex
\begin{abstract}
    %The latent space of diffusion models fundamentally determines their learning efficiency and generation quality. 
    While recent advances in the latent space have driven substantial progress in single-image generation, the optimal latent space for novel view synthesis (NVS) remains largely unexplored. In particular, NVS requires geometrically consistent generation across viewpoints, but existing approaches typically operate in a view-independent VAE latent space. In this paper, we propose \textbf{Geometric Latent Diffusion (GLD)}, a framework that repurposes a geometrically consistent feature space of geometric foundation models as the latent space for multi-view diffusion. We show that the features of the geometric foundation model not only support high-fidelity RGB reconstruction but also encode strong cross-view geometric correspondences, providing a well-suited latent space for NVS. Through experiments, {GLD} outperforms both VAE and RAE on 2D image quality and 3D consistency metrics, accelerating training by more than $\mathbf{4.4\times}$ compared to the VAE latent space. Notably, {GLD} remains competitive with state-of-the-art methods that leverage large-scale text-to-image pretraining, despite training its diffusion model from scratch without such generative pretraining.
%    \vspace{-7pt}
%   \keywords{Novel View Synthesis \and Latent Diffusion \and Geometric Foundation Models}
\end{abstract}

% The latent space of diffusion models fundamentally determines its learning efficiency and generation quality. While recent advances in latent space have driven substantial progress in single image generation, the optimal latent space for novel view synthesis (NVS) remains largely unexplored. NVS requires geometrically consistent generation across viewpoints, but existing approaches typically operate in a latent space without explicit geometric structure, leaving models to implicitly learn cross-view correspondences during training. We hypothesize that this implicit learning incurs a training bottleneck, and that grounding the diffusion process in a geometrically structured latent space can substantially alleviate this bottleneck. To this end, we propose \textbf{Geometric Latent Diffusion (GLD)}, a framework that repurposes the feature space of a geometric foundation model as the latent space for NVS diffusion. We show that the features of a geometric foundation model not only support high-fidelity RGB reconstruction but also encode strong cross-view geometric correspondences, making them well suited as a generative latent space for NVS. Through extensive experiments, \textbf{GLD} consistently outperforms both VAE-based and DINO latent spaces on 2D image quality and 3D consistency metrics, accelerating training by over 3.2$\times$ compared to the VAE latent space. Notably, \textbf{GLD} remains competitive with approaches that leverage large-scale text-to-image pretraining, despite being trained entirely from scratch.

%% file: sec/1_introduction.tex
\section{Introduction}

% \vspace{-10pt}

Diffusion models~\cite{ho2020denoising,song2020score} have become the dominant framework for image synthesis. Transitioning from pixel space to the variational auto-encoder's (VAE) latent space~\cite{rombach2022high,podell2023sdxl,flux2024}, and further to semantically structured representations~\cite{rae,tong2026scaling,shi2025latent}, has shown that the latent space significantly influences generation quality and training efficiency. However, these insights have been drawn exclusively from 2D image generation, and the design of effective latent spaces for \textbf{geometry-aware generation tasks} remains largely unexplored.

Novel view synthesis (NVS), which predicts unseen viewpoints consistent with an underlying 3D scene~\cite{genwarp, moai}, is a representative geometry-aware generation task. Unlike single-image generation, this task requires maintaining coherent spatial structure across views and geometrically plausible completion of occluded regions. While early generative approaches~\cite{cat3d,shi2023mvdream} have demonstrated photorealistic image quality, they often prioritize appearance over geometric consistency, leading to geometrically inconsistent outputs. To address this, recent diffusion-based NVS methods often leverage external geometry conditioning, such as depth-based warping~\cite{genwarp, moai, cao2025mvgenmaster, viewcrafter}. However, these approaches still rely on latent spaces originally designed for single-image synthesis models, such as the 2D VAEs~\cite{rombach2022high}. This raises a fundamental question: \textbf{can we leverage a latent space in which geometric structure is already encoded, rather than injecting or supervising it externally?}

In this work, we propose \textbf{Geometric Latent Diffusion (GLD)} that utilizes the feature space of geometric foundation models~\cite{da3,wang2025vggt,wang2025pi} such as VGGT~\cite{wang2025vggt} or Depth Anything 3~\cite{da3} as a \textbf{latent space for multi-view diffusion models}. We first show that the features of a geometric foundation model support high-fidelity, view-consistent RGB reconstruction, enabling the diffusion process to operate directly on geometry-aware representations for NVS. By training in this geometry-informed latent space, GLD leverages the rich geometric structure, which provides the necessary grounding for generating view-consistent images. Furthermore, since our framework operates natively on the geometric foundation model's features, synthesized latents can be directly decoded into geometric predictions (\eg, depth maps and camera poses) without additional training.

In addition, geometric foundation models typically produce a hierarchy of multi-level features to reconstruct 3D geometries. To ensure computational efficiency, rather than diffusing the entire multi-level features, we identify an optimal boundary layer level for explicit synthesis. Deeper-layer features beyond this boundary are naturally derived by propagating through the frozen backbone, while shallower features are generated via a cascaded scheme to ensure cross-level alignment.

Through extensive experiments on both in-domain and zero-shot benchmarks, GLD achieves superior pixel-level fidelity and cross-view 3D consistency compared with VAE~\cite{rombach2022high} and RAE~\cite{rae} baselines, accelerating training convergence by over $4.4\times$. Although our diffusion model is trained from scratch on small datasets, GLD remains competitive with state-of-the-art methods~\cite{cat3d, cao2025mvgenmaster, lu2025matrix3d, nvcomposer, kwon2025cameo}, fine-tuned from large-scale text-to-image models. Moreover, zero-shot depth and 3D point clouds decoded from synthesized latents exhibit strong global consistency. These results validate that our GLD framework effectively integrates generative modeling with a geometry-informed latent representation.

\label{sec:intro}

%% file: sec/2_relwork.tex
%\vspace{-5pt}
\section{Related Work}
\label{sec:relwork}
%\vspace{-5pt}
\subsubsection{Novel View Synthesis with Diffusion Models.}
Classical geometry-based approaches to novel view synthesis~\cite{mildenhall2021nerf, kerbl20233d} produce photorealistic renderings but require dense multi-view captures and costly per-scene optimization. Recent multi-view diffusion models~\cite{cat3d, genwarp, moai, seva, cao2025mvgenmaster, kong2025causnvs, lu2025matrix3d, viewcrafter} alleviate these constraints by leveraging generative priors to synthesize novel views from sparse inputs. However, these methods operate in pixel or VAE latent spaces that lack cross-view geometric structure, placing a substantial burden on the model to implicitly discover geometric correspondences~\cite{kwon2025cameo}. We instead train multi-view diffusion models in a latent space that already encodes this structure.
\vspace{-10pt}

\subsubsection{Latent Spaces for Diffusion Models.}
Latent diffusion models (LDMs)~\cite{rombach2022high} have advanced image synthesis by operating in a compressed VAE~\cite{vae} latent space, but it lacks rich structural priors. RAE~\cite{rae,tong2026scaling} and SVG~\cite{shi2025latent} show that frozen semantic encoders~\cite{dinov2,siglip2,dinov3} can be paired with lightweight decoders for high-fidelity reconstruction, and that diffusing in this semantic space yields faster convergence and improved generation quality. However, these advances target single-image generation, leaving open the question of how to design latent spaces for geometry-aware generation tasks. Recent works address this by training dedicated autoencoders that jointly encode appearance and geometry, for single-image~\cite{krishnan2025orchid} and text-to-3D~\cite{yang2025prometheus} generation. Our work instead repurposes the feature space of an existing geometric foundation model~\cite{da3} as the latent space for diffusion, providing the model with cross-view geometric priors.

% Latent diffusion models (LDMs)~\cite{rombach2022high} have advanced image synthesis by operating in a compressed VAE~\cite{vae} latent space. Because the VAE is optimized primarily for photometric reconstruction, it lacks rich structural priors.  RAE~\cite{rae,tong2026scaling} and SVG~\cite{shi2025latent} show that frozen semantic encoders~\cite{dinov2,siglip2} can be paired with lightweight decoders for high-fidelity image reconstruction. Furthermore, training diffusion models in this semantic space yields faster convergence and improved generation quality compared to training in VAE latent spaces. However, these advances target single-image generation, leaving open the question of how to design latent spaces for geometry-aware generation tasks. Recent works address this by training dedicated autoencoders that jointly encode appearance and geometry, such as Orchid~\cite{krishnan2025orchid} for single-image generation and Prometheus~\cite{yang2025prometheus} for text-to-3D generation. Our work instead repurposes the feature space of an existing geometric foundation model~\cite{da3} as the latent space for multi-view diffusion. Since these features encode cross-view geometric representations, they provide the diffusion model with informative geometric priors and enable zero-shot geometric predictions without additional training.

\vspace{-10pt}

\subsubsection{Geometric Foundation Models.}
Geometric foundation models have introduced a paradigm shift in 3D vision, moving from optimization-based feature matching~\cite{schonberger2016structure} to purely feed-forward scene understanding. Building on the pairwise formulation of DUSt3R~\cite{wang2024dust3r}, recent models~\cite{da3, wang2025vggt, wang2025pi, keetha2025mapanything} have enabled feed-forward dense 3D reconstruction from arbitrary unposed views, jointly predicting camera parameters and depth maps. While recent analyses reveal that the internal representations of these networks encode strong geometric correspondences~\cite{han2025emergent}, their utility has been largely limited to discriminative tasks. We bridge this gap by showing that the feature space of a geometric foundation model~\cite{da3} can serve as an effective latent space for novel view synthesis.

%% file: sec/3_preliminary.tex
%\vspace{-7pt}
\section{Preliminaries}
\label{sec:preliminaries}
\vspace{-5pt}
\subsubsection{Representation Autoencoder.}
Representation Autoencoder (RAE)~\cite{rae,tong2026scaling} replaces the conventional VAE~\cite{vae} in latent diffusion~\cite{rombach2022high} with a pretrained, frozen vision encoder~\cite{dinov2,siglip2} $\mathcal{E}(\cdot)$ and a trainable decoder $\mathcal{D}(\cdot)$, directly adopting the encoder's feature space as the diffusion latent space.

Specifically, the decoder is trained to reconstruct RGB images from features in this representation space, showing that these features are not only semantically rich but also sufficient for high-fidelity reconstruction. Formally, given a single-view image $\mathcal{I} \in \mathbb{R}^{H \times W \times 3}$, where $H$ and $W$ denote the height and width, respectively, the encoder extracts a tokenized feature representation
\begin{equation}\mathbf{F} = \mathcal{E}(\mathcal{I}) \in \mathbb{R}^{T \times C},\end{equation}
where $T$ is the token sequence length and $C$ is the channel dimension. RAE further shows that diffusion models can be trained directly in this representation space, yielding faster convergence and stronger generative performance than training in conventional VAE latent spaces~\cite{vae}. During generation, the diffusion model synthesizes $\tilde{\mathbf{F}}$, and the synthesized image $\tilde{\mathcal{I}}$ is then obtained by decoding the synthesized feature: $\tilde{\mathcal{I}} = \mathcal{D}(\tilde{\mathbf{F}})$.

\vspace{-10pt}

\subsubsection{Geometric Foundation Models.}
Recent foundation models for geometry~\cite{da3,wang2025vggt,wang2025pi} typically consist of a Vision Transformer (ViT)~\cite{dosovitskiy2020image} encoder $\mathcal{E}_{\text{geo}}(\cdot)$ and a DPT-based geometric decoder $\mathcal{D}_{\text{geo}}(\cdot)$. To process multi-view inputs, these architectures often incorporate 3D attention in addition to standard intra-image self-attention, which enables joint reasoning across multiple frames. Given multi-view images $\mathbf{I} \in \mathbb{R}^{V \times H \times W \times 3}$, where $V$ is the number of input views, the encoder extracts multi-view feature sequences at $L$ levels (often $L=4$):
\begin{equation}\{{\mathbf{F}_l}\}_{l=0}^{L-1} = \mathcal{E}_{\text{geo}}(\mathbf{I}),
\end{equation} 
where $\mathbf{F}_l \in \mathbb{R}^{V\times T \times C}$ denotes the multi-view feature sequence at level $l$, with $T$ the token sequence length and $C$ the channel dimension. The geometric decoder then aggregates these multi-level features to produce dense geometric predictions, such as depth or camera parameters:\begin{equation}\mathbf{G} = \mathcal{D}_{\text{geo}}({\{\mathbf{F}_l}\}_{l=0}^{L-1}),
\end{equation}
where $\mathbf{G}$ denotes a set of geometric predictions for each input view.

%% file: sec/5_method.tex
%\vspace{-7pt}
\section{Method}\label{sec:method}
\input{fig/architecture}

\subsection{Overview}
\label{sec:overview}
Our goal is to harness the feature space of geometric foundation models~\cite{da3,wang2025vggt,wang2025pi} as the latent space for multi-view diffusion, enabling high-fidelity novel view synthesis (NVS). Specifically, we adopt the Depth Anything 3 (DA3)~\cite{da3} as our primary backbone, which extracts features across $L=4$ intermediate levels. We also explore VGGT~\cite{wang2025vggt} as an additional backbone in Appendix~\ref{app:vggt}. Given a set of $N$ source images $\mathbf{I}^\text{src} \in \mathbb{R}^{N \times H \times W \times 3}$ with camera poses $\mathbf{P}^\text{src}$, and $M$ target camera poses $\mathbf{P}^\text{tgt}$, we seek to synthesize the corresponding target views $\tilde{\mathbf{I}}^\text{tgt} \in \mathbb{R}^{M \times H \times W \times 3}$. Rather than operating directly in pixel space or VAE space~\cite{rombach2022high}, our framework generates multi-view, multi-level features from geometric foundation models~\cite{da3,wang2025vggt,wang2025pi}, which are subsequently decoded into the target views.

    Because the geometric foundation model's 3D attention jointly encodes source and target views, the resulting features are inherently coupled. We therefore generate both source and target features across all levels, denoted as $\{\tilde{\mathbf{F}}_l^\text{src}\}_{l=0}^{L-1}$ with $\tilde{\mathbf{F}}_l^\text{src} \in \mathbb{R}^{N\times T \times C}$, and $\{\tilde{\mathbf{F}}_l^\text{tgt}\}_{l=0}^{L-1}$ with $\tilde{\mathbf{F}}_l^\text{tgt} \in \mathbb{R}^{M\times T \times C}$. At each level $l$, the joint feature $\tilde{\mathbf{F}}_l \in \mathbb{R}^{(N+M)\times T \times C}$ is formed by concatenating the source and target features along the view dimension, yielding $\tilde{\mathbf{F}}_l = [\tilde{\mathbf{F}}_l^\text{src}, \tilde{\mathbf{F}}_l^\text{tgt}]$ with concatenation operator $[\cdot,\cdot]$. Finally, a dedicated RGB decoder $\mathcal{D}_{\text{rgb}}(\cdot)$ maps the complete synthesized feature set back to the pixel space to render the target views via $\tilde{\mathbf{I}}^\text{tgt} = \mathcal{D}_{\text{rgb}}(\{\tilde{\mathbf{F}}^\text{tgt}_l\}_{l=0}^{L-1})$. The target geometry $\tilde{\mathbf{G}}^\text{tgt}$ is also decoded such that $\tilde{\mathbf{G}}^\text{tgt} = \mathcal{D}_{\text{geo}}(\{\tilde{\mathbf{F}}^\text{tgt}_l\}_{l=0}^{L-1})$.

To this end, as illustrated in \cref{fig:architecture}, \textbf{Geometric Latent Diffusion (GLD)} framework employs a three-stage pipeline. First, \S~\ref{sec:reconstruction} validates the reconstruction capacity of the geometric feature space by training $\mathcal{D}_{\text{rgb}}(\cdot)$ to decode multi-level features into RGB images. Second, to avoid the substantial cost of diffusing all $L$ feature levels, \S~\ref{sec:select_level} identifies the optimal boundary layer $k$. Since deeper features can be obtained by propagating the boundary feature $\tilde{\mathbf{F}}_k$ through $\mathcal{E}_{\text{geo}}(\cdot)$, we only require explicit synthesis up to this boundary feature. Finally, \S~\ref{sec:cascade} employs a cascaded scheme to synthesize shallower features from $\tilde{\mathbf{F}}_k$ to ensure cross-level alignment.

\vspace{-10pt}

\subsection{Validating the Reconstruction Capability of Geometric Features}\label{sec:reconstruction} 

To validate the suitability of DA3's feature space for generative modeling, we first verify that its features can be decoded into high-fidelity images. We train a ViT-based decoder $\mathcal{D}_{\text{rgb}}(\cdot)$ to reconstruct RGB images from the multi-level features $\{\mathbf{F}_l\}_{l=0}^{L-1}$ extracted by the frozen encoder $\mathcal{E}_{\text{geo}}(\cdot)$. To ensure $\mathcal{D}_{\text{rgb}}(\cdot)$ effectively leverages the full signal, we introduce a level-wise dropout strategy during training. By randomly masking individual levels in $\{\mathbf{F}_l\}_{l=0}^{L-1}$, we force the decoder to reconstruct from partial inputs, improving its robustness. 

\input{table/recon_ours}
As shown in \cref{tab:recon_comp} and \cref{fig:recon_qualitative}, $\mathcal{D}_{\text{rgb}}(\cdot)$ successfully recovers the input images with high fidelity while preserving fine-grained details. These results show that the DA3 feature space is suitable as the latent space for our diffusion process. Further details and comparison with other baselines are available in \S~\ref{sec:decoder}.

\vspace{-10pt}

\subsection{Multi-view Diffusion and Determining the Boundary Layer}\label{sec:select_level}

While the DA3 feature space provides a sufficiently expressive latent for high-fidelity image reconstruction, explicitly synthesizing the full multi-level set $\{\mathbf{F}_l\}_{l=0}^3$ is computationally prohibitive. Since deeper features ($l > k$) can be deterministically derived by propagating a shallower feature $\tilde{\mathbf{F}}_k$ through the frozen layers of $\mathcal{E}_{\text{geo}}$, we only require explicit synthesis up to an optimal boundary $k$. To identify this boundary, we first train four independent diffusion models $\{\mathcal{M}_l\}_{l=0}^3$, each dedicated to synthesizing the target feature at a specific level $l$. We then perform a comparative evaluation by varying the synthesis boundary $k \in \{0,1,2,3\}$ to identify the shallowest boundary sufficient for high-quality NVS.

\vspace{-10pt}

\subsubsection{Multi-view Diffusion Architecture and Training.}
We adopt the $\text{DiT}^{\text{DH}}$~\cite{ddt}
architecture from RAE~\cite{rae} and train it with a flow-matching
objective~\cite{lipman2023flow} to synthesize the joint feature map
$\tilde{\mathbf{F}}_l$ for all $V = N + M$ views.
As illustrated in \cref{fig:architecture}, we incorporate
3D self-attention~\cite{cat3d} with PRoPE~\cite{prope} and condition on
Pl\"ucker ray embeddings to enforce geometric consistency across views.

Each model $\mathcal{M}_l$ is conditioned on the source-only features
$\mathbf{F}_l^\text{src}$, extracted by the frozen DA3 encoder from the $N$ source images alone, by concatenating them with the noisy latent
$\mathbf{z}_t$ along the channel dimension. Note that $\mathbf{F}_l^\text{src}$ is extracted without access to target views, whereas the source portion of the full joint feature $\mathbf{F}_l$ is influenced by 3D attention over all $V$ views. Because the decoder and downstream stages require features from all views, we design the model to jointly generate $\mathbf{F}_l$ rather than generate only the target views.

\vspace{-15pt}

\subsubsection{Boundary Layer Evaluation.}
\label{sec:boundary}
To identify the optimal boundary, we assess how each level $k$ contributes to the generation by providing the decoder $\mathcal{D}_{\text{rgb}}$ with a complete multi-level set $\{\tilde{\mathbf{F}}_l\}_{l=0}^3$. For a given boundary $k$, we explicitly synthesize the features up to that level ($l\le k$), $\{{\tilde{\mathbf{F}}}_l\}_{l=0}^{k}$, using the corresponding set of independently trained models $\{\mathcal{M}_l\}_{l=0}^{k}$. The remaining deeper levels ($l > k$) are then deterministically derived by passing $\tilde{\mathbf{F}}_k$ through the frozen layers of $\mathcal{E}_{\text{geo}}(\cdot)$.

\input{table/level_selection_view2}

As shown in \cref{tab:level_abl}, synthesizing up to level 1 achieves superior NVS performance. Shifting the boundary from level 0 to level 1 improves both RGB quality and geometric accuracy, suggesting that level 1 provides a more effective latent representation for synthesis. Conversely, using deeper levels ($2$ or $3$) as the boundary leads to a consistent degradation in metrics, likely due to the loss of fine-grained spatial details in abstract feature spaces. Consequently, we fix level 1 as the synthesis boundary $k$ for our full framework, as illustrated in \cref{fig:architecture}(a). Further analysis of this selection is provided in \S~\ref{sec:level_discussion}.

\vspace{-10pt}

\subsection{Cascaded Feature Generation}\label{sec:cascade}
Based on the evaluation in \S~\ref{sec:boundary}, we fix the synthesis boundary at level 1 and generate the corresponding multi-level set. While level 0 can be synthesized independently, generating them separately causes misalignment across the feature hierarchy. To provide the coherent input required by the decoder, we instead employ a cascaded model $\mathcal{M}_{1\rightarrow0}$ that synthesizes level 0 conditioned on the generated level 1 latent, which is illustrated in \cref{fig:architecture}(b). $\mathcal{M}_{1\rightarrow0}$ shares the same architecture and training configuration as $\mathcal{M}_0$.

To handle the imperfect latents encountered during inference, we train $\mathcal{M}_{1\rightarrow0}$ by conditioning on a noisy version of the ground-truth $\mathbf{F}_1$. This strategy improves the model's robustness and ensures $\tilde{\mathbf{F}}_0$ is anchored to $\tilde{\mathbf{F}}_1$, providing the alignment the decoder requires. A quantitative validation of this cascaded approach over independent generation is provided in \S~\ref{sec:ablation}.

%% file: fig/architecture.tex
\begin{figure}[t]
  \centering
  \includegraphics[width=\linewidth]{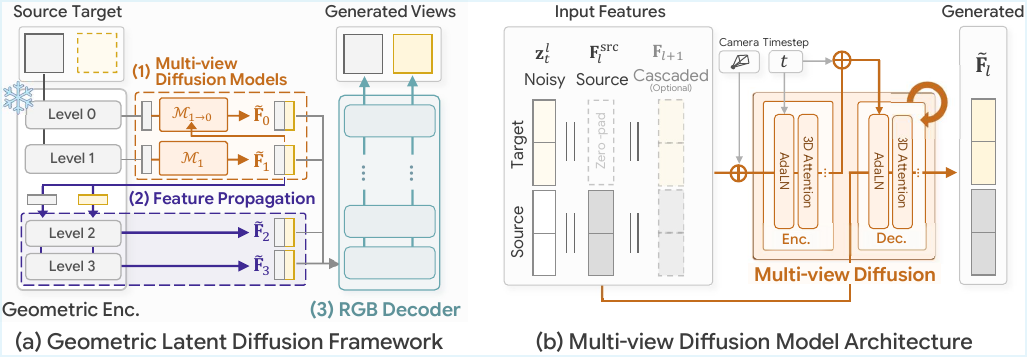}
  \vspace{-15pt}
  \caption{\textbf{Overview of the proposed framework.}
  \textbf{(a)} Our NVS framework consists of three stages:
  \textbf{{\color[HTML]{BD5C03}(1) multi-view diffusion models}} synthesize features up to boundary layer $k{=}1$ via cascaded generation,
  \textbf{{\color[HTML]{2A1187}(2) feature propagation}} derives deeper features through the frozen DA3 encoder, and
  \textbf{{\color[HTML]{5C9FA8}(3) the RGB decoder}} maps the full multi-level features to target views.
  \textbf{(b)} The detailed architecture of a multi-view diffusion model.}
  \label{fig:architecture}
  \vspace{-12pt}
\end{figure}

%% file: table/recon_ours.tex
\begin{figure}[t]
  \centering
  \begin{minipage}[t]{0.6\linewidth}
    \vspace{0pt}
    \centering
    \includegraphics[width=\linewidth]{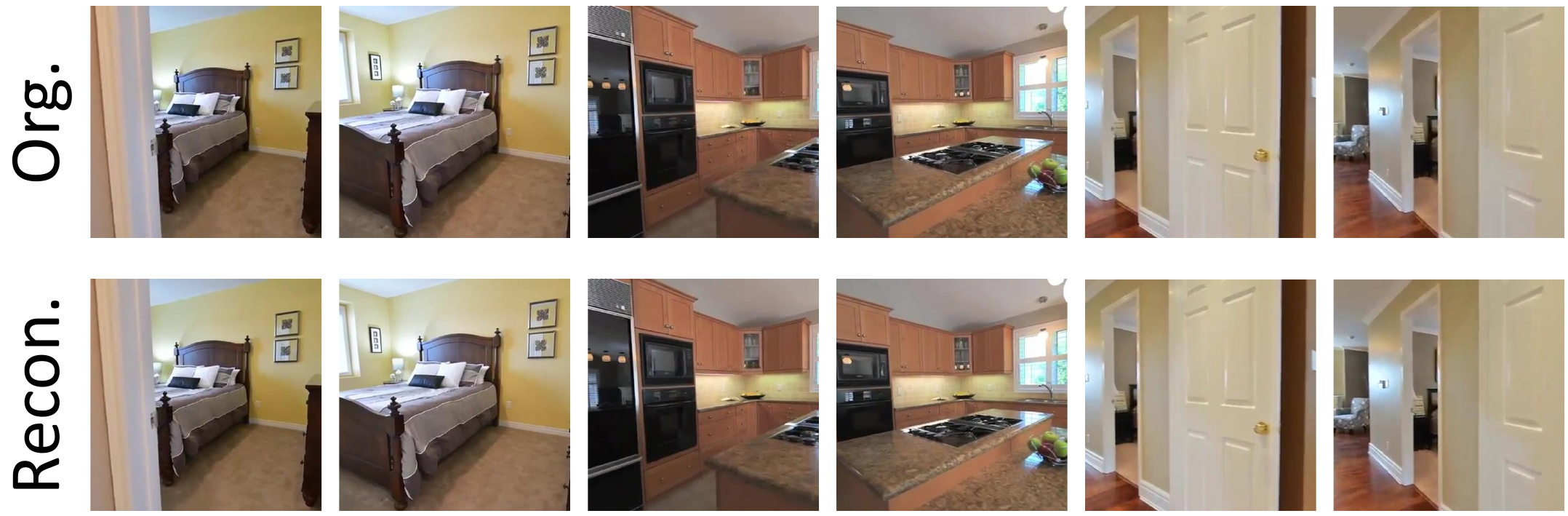}
    \vspace{-15pt}
    \captionof{figure}{\textbf{Reconstruction Results.} We visualize the reconstructed images from $\mathcal{D}_\text{rgb}(\cdot)$.}
    \label{fig:recon_qualitative}
  \end{minipage}
  \hfill
  \begin{minipage}[t]{0.36\linewidth}
    \vspace{0pt}
    \vspace{-\abovecaptionskip}
    \centering
    \captionof{table}{\textbf{Reconstruction Fidelity.} We evaluate our RGB decoder on 4,000 samples from the Re10K~\cite{re10k} test set.}
    \label{tab:recon_comp}
    \vspace{2pt}
    \setlength{\tabcolsep}{9pt}
     \resizebox{0.88\linewidth}{!}{%
      \begin{tabular}{lc}
      \toprule
      Metrics & Decoder ($\mathcal{D}_{\text{rgb}}$) \\
      \midrule
      PSNR $\uparrow$  & 35.41 \\
      SSIM $\uparrow$  & 0.960   \\
      LPIPS $\downarrow$ & 0.019 \\
      \bottomrule
    \end{tabular}%
  }
  \end{minipage}
  \vspace{-20pt}
\end{figure}

% \begin{wraptable}[13]{r}{0.5\textwidth}
%   \centering
%   \vspace{-25pt}
%   % Figure Section
%   \includegraphics[width=\linewidth]{pdf/recon_ours.pdf}
%   \vspace{-20pt}
%   \captionof{figure}{\textbf{Reconstruction visualization.} We visualization the reconstruction results from our RGB decoder.}
%   \label{fig:recon_qualitative}
%   % Table Section
%   \vspace{-7pt}
%   \captionof{table}{\textbf{Reconstruction Fidelity.} We evaluate our RGB decoder on 4,000 samples from Re10K~\cite{re10k} test set.}
%   \label{tab:recon_comparison}
%   \setlength{\tabcolsep}{10pt}
%   \resizebox{0.7\linewidth}{!}{%
%     \begin{tabular}{lc}
%       \toprule
%       \textbf{Metric} & \textbf{Decoder ($\mathcal{E}_{\text{rgb}}$)} \\
%       \midrule
%       PSNR $\uparrow$  & 35.38 \\
%       SSIM $\uparrow$  & ---   \\
%       LPIPS $\downarrow$ & 0.019 \\
%       \bottomrule
%     \end{tabular}%
%   }
%   \vspace{-15pt}
% \end{wraptable}

%% file: table/level_selection_view2.tex
\begin{table}[t]
    \centering
    \caption{\textbf{Boundary Selection.} We evaluate NVS performance by varying the synthesis boundary $k$. Explicitly synthesizing up to level $l=1$ (Boundary $k=1$) using $\mathcal{M}_0$ and $\mathcal{M}_1$ provides the best performance across all metrics.}
    \label{tab:level_abl}
    \vspace{-5pt}
    \resizebox{\columnwidth}{!}{%
        \begingroup
        \setlength{\tabcolsep}{8pt} 
        \begin{tabular}{@{}ll|ccc|ccc@{}}
            \toprule
            \multicolumn{2}{c|}{Synthesis Configuration} & \multicolumn{3}{c|}{RGB Metrics} & \multicolumn{3}{c}{Depth Metrics} \\
            Boundary $k$ & Models $\{\mathcal{M}_l\}_{l=0}^k$ & PSNR $\uparrow$ & SSIM $\uparrow$ & LPIPS $\downarrow$ & AbsRel $\downarrow$ & RMSE $\downarrow$ & $\delta < 1.25 \uparrow$ \\ \midrule
            $k=0$ & $\{\mathcal{M}_0\}$ & 12.55 & 0.323 & 0.579 & 0.267 & 0.400 & 0.641 \\
            \rowcolor{lightpurblue}
            \textbf{$k=1$} & \textbf{$\{\mathcal{M}_0, \mathcal{M}_1\}$} & \textbf{13.61} & \textbf{0.366} & \textbf{0.555} & \textbf{0.191} & \textbf{0.311} & \textbf{0.744} \\
            $k=2$ & $\{\mathcal{M}_0,\mathcal{M}_1,\mathcal{M}_2\}$ & 13.35 & 0.355 & 0.566 & 0.254 & 0.393 & 0.659 \\
            $k=3$ & $\{\mathcal{M}_0,\mathcal{M}_1,\mathcal{M}_2,\mathcal{M}_3\}$ & 13.35 & 0.355 & 0.567 & 0.260 & 0.402 & 0.647 \\ \bottomrule
        \end{tabular}%
        \endgroup
    }
    \vspace{-20pt}
\end{table}

%% file: sec/6_experiment.tex
\input{table/main_table}
%\vspace{-10pt}
\section{Experiments}
\subsection{Setup} 
\subsubsection{Datasets.}
GLD is trained from scratch on four datasets, RealEstate10K~\cite{re10k} (Re10K), DL3DV~\cite{ling2024dl3dv}, HyperSim~\cite{hypersim}, and TartanAir~\cite{tartanair2020iros}. Each training sample consists of $V=8$ views, where 1 to 4 views are randomly selected as source views while the rest are masked as targets. For the main evaluation, we evaluate on two in-domain benchmarks, Re10K and DL3DV, which overlap with the training distribution, and on one out-of-domain object-centric benchmark, Mip-NeRF 360~\cite{mipnerf}, to evaluate generalization to unseen scene types. We use $N=2$ source views and measure performance on the target views across $200$ samples per dataset. Additional details are provided in Appendix~\ref{app:dataset}.

\vspace{-10pt}

\subsubsection{Training Details.} Our diffusion model operates in the latent space of DA3-Base~\cite{da3}. We train the diffusion model using AdamW~\cite{loshchilov2017decoupled} with a fixed learning rate of $5{\times}10^{-5}$, a batch size of 48, and EMA decay of 0.9995. We apply 10\% dropout to camera embeddings for classifier-free guidance~\cite{ho2022classifier}, with a CFG scale of $1.5$ at inference. For each sample, the training resolution is randomly chosen from ${(504{\times}504),(504{\times}378),(504{\times}336)}$, and $(504{\times}280)$, matching the set of resolutions used to train DA3. GLD is trained on 8 B200 GPUs for 175k iterations. Additional details are provided in Appendix~\ref{app:training}.
% Decoder학습 디테일

\vspace{-10pt}

\subsubsection{Baselines.}
We compare GLD against two categories of baselines. The first category comprises general-purpose visual encoders, such as VAE~\cite{rombach2022high} and DINO~\cite{dinov2}, to assess whether the latent space of DA3~\cite{da3} is better suited for novel view synthesis than general-purpose visual representations. We additionally evaluate VGGT~\cite{wang2025vggt} as an alternative geometric foundation model backbone to further examine the effectiveness of geometry-aware latent spaces for NVS; these results are provided in Appendix~\ref{app:vggt}. For VAE, we use the Stable Diffusion encoder-decoder~\cite{rombach2022high}. For DINO, we adopt DINOv2 ViT-B/14 with registers~\cite{dinov2,dinoregister} as the encoder and train a decoder from scratch. In all cases, the diffusion model is trained from scratch for the same number of iterations using the same architecture as GLD. Additional implementation details are provided in the Appendix~\ref{app:architecture}.

The second category comprises state-of-the-art diffusion-based NVS methods, including MVGenMaster~\cite{cao2025mvgenmaster}, Matrix3D~\cite{lu2025matrix3d}, CAMEO~\cite{kwon2025cameo}, NVComposer~\cite{nvcomposer}, and CAT3D$^\dagger$~\cite{cat3d}\footnote{Since the official implementation of CAT3D is unavailable, we use the model and checkpoint reproduced in CAMEO~\cite{kwon2025cameo}.}. These models typically leverage powerful generative priors by fine-tuning from large-scale pre-trained weights. Note that CAMEO and CAT3D$^\dagger$ are trained exclusively on the Re10K dataset, whereas the remaining methods incorporate both scene-centric and object-centric data during training.

\vspace{-10pt}

\subsubsection{Evaluation Protocol.}
We evaluate the 2D image fidelity of generated target views using standard NVS metrics: PSNR, SSIM, and LPIPS. To assess the 3D geometric consistency of generated views, we further incorporate camera estimation errors, reprojection error~\cite{du2026videogpa}, and MEt3R~\cite{asim25met3r}. Specifically, for camera errors, we extract camera poses from the generated views using an external estimator~\cite{wang2025vggt} to compute the Absolute Trajectory Error (ATE), and Relative Pose Errors for rotation (RPE$_r$) and translation (RPE$_t$). These camera errors explicitly evaluate condition fidelity by measuring how accurately the generated images adhere to the target pose conditioning. Furthermore, reprojection error and MEt3R quantify the underlying 3D geometric consistency across the generated images. Reprojection error measures the spatial re-alignment accuracy of reconstructed 3D points, while MEt3R evaluates multi-view consistency using projected feature similarity.

\input{fig/main_qual}

\vspace{-10pt}

\subsection{Quantitative Results} 
\subsubsection{2D Metrics.}
We evaluate image synthesis quality on two in-domain datasets (Re10K and DL3DV) and one zero-shot, out-of-domain benchmark (Mip-NeRF 360). First, we compare our performance against VAE and DINO encoder baselines. As shown in \cref{tab:comparison}, our method consistently outperforms both baselines in PSNR, SSIM, and LPIPS across all benchmarks. The consistent gains confirm that the DA3 feature space provides a more suitable latent representation for novel view synthesis (NVS) than general-purpose visual encoders.

Second, we evaluate GLD against state-of-the-art NVS methods that leverage massive diffusion priors via large-scale text-to-image (T2I) pretraining. Despite being trained from scratch on smaller datasets, GLD surpasses all baselines across all 2D metrics on both in-domain benchmarks. For the out-of-domain evaluation, which consists mainly of object-centric samples, GLD still achieves state-of-the-art PSNR and highly competitive results across the remaining 2D metrics. This generalization is particularly notable given that GLD is trained exclusively on scene-level data, whereas competing baselines~\cite{cao2025mvgenmaster, kwon2025cameo, nvcomposer} incorporate object-centric datasets~\cite{reizenstein2021common, deitke2023objaverse} during fine-tuning.

\vspace{-10pt}

\subsubsection{3D Metrics.}
We next evaluate cross-view geometric consistency. As shown in \cref{tab:comparison}, GLD consistently outperforms the VAE and DINO baselines across most 3D metrics on every benchmark. The most substantial gains are in pose accuracy, where GLD achieves up to a $2.8\times$ lower ATE and a $2.6\times$ lower RPE compared to the baselines. Reprojection error and MEt3R also show consistent improvements across most evaluation settings. These significant improvements compared to VAE and DINO latent spaces demonstrate that GLD yields superior 3D consistency and notably accurate adherence to the target pose.

 % (Matrix3D, CAT3D$^\dagger$, CAMEO, and NVComposer)
 
When compared to most fine-tuned methods, GLD exhibits significantly superior 3D consistency across in-domain benchmarks. This performance gap highlights a fundamental limitation of relying solely on generic T2I priors for NVS. Compared to MVGenMaster, GLD achieves superior performance on DL3DV, while remaining highly competitive on Re10K. Note that MVGenMaster utilizes an external depth estimator~\cite{depth-pro} and warps source RGB and depth to the target view as condition inputs, making it heavily reliant on explicit, dense geometric priors. In contrast, GLD operates in a multi-view geometry-aware latent space without requiring explicit geometry conditions (\eg, depth maps) during inference. Although GLD achieves slightly lower pose accuracy on the out-of-domain benchmark than Matrix3D and MVGenMaster, this is expected, as those baselines benefit from strong T2I priors and are additionally fine-tuned on object-centric datasets. Despite relying exclusively on scene-level training without external warping, GLD still achieves the lowest reprojection error and a competitive MEt3R score. This shows that our DA3 latent space effectively encodes the geometric structures necessary for robust, coherent multi-view NVS.

\input{table/v_ablation}

\vspace{-10pt}

\subsection{Qualitative Results}
\cref{fig:main_qual} presents qualitative comparisons of generated target views. The left column (a) shows methods trained from scratch, while the right column (b) shows methods fine-tuned from pretrained T2I models. GLD produces structurally coherent views that closely follow the ground-truth layout, outperforming the VAE and DINOv2 baselines particularly under large viewpoint changes, and remaining competitive with fine-tuned methods despite not leveraging T2I pretraining. We note that the method that relies on warped source views from an external geometry estimator~\cite{cao2025mvgenmaster} produces sharp outputs but can exhibit noticeable artifacts when the estimation fails, as seen in the fourth column of the first scene, whereas GLD avoids such failure modes. Additional qualitative results are provided in Appendix~\ref{app:additional_results}.

\vspace{-10pt}

\subsection{Decoder Performance}\label{sec:decoder}
\input{fig/level_analysis}

We train the ViT-based RGB decoder, $\mathcal{D}_{\text{rgb}}$, on a combined dataset of Re10K~\cite{re10k} and DL3DV~\cite{ling2024dl3dv}. Following RAE~\cite{rae}, the model is optimized using a weighted sum of $\mathcal{L}_1$, LPIPS, and adversarial losses. We utilize the AdamW optimizer and a cosine learning rate scheduler with a 1-epoch warmup.

The decoder is trained using a multi-resolution strategy involving resolutions of $(504, 504)$, $(504, 378)$, $(504, 336)$, and $(504, 280)$. We employ a global batch size of 128 and an EMA decay of 0.9978. The entire training process is conducted on 8 NVIDIA B200 GPUs, spanning approximately 170k steps. For further comparison, we additionally train a baseline RAE decoder on DINOv2~\cite{dinov2} features using the same architecture and training configuration.

\input{table/reconstruction}

In \cref{tab:recon_comparison}, we compare $\mathcal{D}_{\text{rgb}}$ with the standard pretrained VAEs used in Stable Diffusion~\cite{rombach2022high} and SDXL~\cite{podell2023sdxl}. Our decoder reconstructs high-fidelity images from DA3 features and performs competitively with these widely used models. These results indicate that multi-level DA3 features form a sufficiently expressive latent space for high-quality image synthesis.
% In Tab.~\ref{tab:recon_comparison}, we compare $\mathcal{D}_{\text{rgb}}$ with two widely used VAEs~\cite{podell2023sdxl,rombach2022high}.  our decoder reconstructs high-fidelity images from DA3 features and outperforms the used in Stable Diffusion~\cite{rombach2022high} and SDXL~\cite{podell2023sdxl}. These results indicate that multi-level DA3 features form a sufficiently expressive latent space for high-quality image reconstruction.

\vspace{-10pt}

\subsection{Discussion on Boundary Layer Selection}\label{sec:level_discussion}
In \S~\ref{sec:select_level}, we found that boundary layer $k=1$ yields the best NVS performance. To understand why, we analyze the four levels of features (levels 0 to 3). Specifically, we examine features from geometric and photometric perspectives by measuring their cross-view correspondence and their capacity for high-fidelity image reconstruction, respectively.

For geometric correspondence, we follow the evaluation protocol in Probe3D~\cite{el2024probing} and assess features on the ScanNet~\cite{dai2017scannet} dataset using PCK, the fraction of points that are correctly matched within a distance threshold. Our results in \cref{tab:pck_eval} show that level 1 and level 2 achieve high PCK scores, even outperforming DINOv2~\cite{dinov2}. This indicates that these levels encode stable 3D structures. In contrast, level 0 exhibits poor correspondence; being the shallowest layer, it does not encode the complex geometric relationships needed for cross-view matching.

To evaluate photometric information, we utilize the RGB decoder $\mathcal{D}_{rgb}$ from \S~\ref{sec:reconstruction} to reconstruct the original image. Because $\mathcal{D}_{rgb}$ is trained with layer dropout, it is capable of reconstructing images from a single feature level. By reconstructing from each level independently and measuring the error, we find that levels 0 and 1 retain rich appearance cues, producing accurate colors and textures. However, deeper features, such as level 2, discard essential photometric details. This results in noticeable color loss and smoothed textures (\cref{fig:recon_vis}), as well as a notable drop in PSNR compared to shallower levels (\cref{tab:recon_eval}).

These results indicate that level 1 successfully meets both requirements. It maintains a high degree of geometric alignment comparable to level 2, while retaining substantially more photometric information than the deeper features. This observation provides a straightforward explanation for why level 1 serves as the optimal latent space for our NVS framework. 

\vspace{-10pt}

\subsection{Ablation Studies}
\label{sec:ablation}
\vspace{-2pt}
\subsubsection{Varying the Number of Source Images.}

In this section, we evaluate the robustness of GLD under varying numbers of source views by comparing against the VAE~\cite{rombach2022high} and DINOv2~\cite{dinov2} baselines. Following the evaluation protocol from our main experiments, we present the quantitative results in \cref{tab:cond_ablation}. Notably, the margin of improvement in 3D metrics grows as the number of source views decreases. For instance, on DL3DV with $N{=}1$, GLD achieves over $3\times$ lower ATE and RPE$_r$ compared to both baselines, whereas with $N{=}4$ the gap is approximately $2\times$. This trend is consistent across both datasets, suggesting that the geometric priors in the DA3 feature space become particularly beneficial when fewer visual cues are available.

\vspace{-10pt}

\subsubsection{Effectiveness of the Cascaded Design.}
\input{table/independent}

In \S~\ref{sec:cascade}, we generate $\tilde{\mathbf{F}}_0$ by conditioning on the previously synthesized $\tilde{\mathbf{F}}_1$ to ensure consistency across feature levels. To validate this design, we compare against an independent baseline where $\mathcal{M}_0$ and $\mathcal{M}_1$ generate their respective features without seeing each other's output. As shown in \cref{tab:cascading}, evaluated on Re10K with $N{=}4$ source views, the cascading approach consistently improves both 2D and 3D metrics, confirming that allowing shallower features to be conditioned by deeper ones leads to more coherent multi-level representations.

\vspace{-10pt}

\subsection{Geometry Evaluation}\label{sec:3d_reconstruction}

\input{table/depth}

Since GLD generates the multi-level DA3 features, we can leverage the original, pretrained DPT-based decoder $\mathcal{D}_{\text{DA3}}$ to predict ray and depth maps without any additional training or fine-tuning. This allows us to obtain dense geometric predictions as a zero-shot byproduct of the generation process. We evaluate the fidelity of these maps through depth metrics and 3D point-cloud visualizations.

\vspace{-10pt}

\subsubsection{Depth Evaluation.}

We compare our generated depth maps against Matrix3D~\cite{lu2025matrix3d}, which jointly generates depth and RGB images. We evaluate performance on the ETH3D dataset~\cite{eth}, which provides high-quality ground-truth depth maps. As shown in \cref{tab:eth3d_cond2}, our model produces more accurate depth estimates, suggesting that operating in the DA3 latent space provides effective geometry grounding.

\vspace{-10pt}

\input{fig/3d_vis}

\subsubsection{3D Reconstruction Visualization.}
We visualize the 3D reconstruction by unprojecting the synthesized pixels into 3D space using the generated depth and ray maps. As illustrated in \cref{fig:reconstruction_vis}, the resulting point clouds exhibit consistent 3D geometry across diverse camera trajectories, confirming that our generated latent features are geometrically coherent with the underlying scene. We further compare against Matrix3D~\cite{lu2025matrix3d} in \cref{fig:3d_compare}. While Matrix3D produces noticeable geometric distortions, GLD yields significantly cleaner and more coherent reconstructions, consistent with the quantitative results in \cref{tab:eth3d_cond2}.
\label{sec:experiment}

%% file: table/main_table.tex
\begin{table*}[t]
\centering
\caption{Quantitative comparison of different methods across in-domain~\cite{re10k, ling2024dl3dv} and out-of-domain~\cite{mipnerf} datasets on 2D and 3D metrics. \textbf{Bold} and \underline{underlined} values indicate the best and second-best results, respectively. We use reproduced CAT3D$^\dagger$~\cite{kwon2025cameo}. }
\label{tab:comparison}
\vspace{-5pt}
\setlength{\tabcolsep}{7pt}    % tighter column padding (default 6pt)
\small
\renewcommand{\arraystretch}{0.85} % tighter row spacing (default 1.0)
    \resizebox{\textwidth}{!}{%
        \begin{tabular}{l|c|ccc|ccccc}
        \toprule
        \multirow{2}{*}{Methods} & \multirow{2}{*}[-2pt]{\shortstack{From\\Scratch}} & \multicolumn{3}{c|}{2D Metrics} & \multicolumn{5}{c}{3D Metrics} \\
        \cmidrule(lr){3-5} \cmidrule(lr){6-10}
         & & PSNR $\uparrow$ & SSIM $\uparrow$ & LPIPS $\downarrow$ & ATE $\downarrow$ & RPE$_r$ $\downarrow$ & RPE$_t$ $\downarrow$ & Reproj $\downarrow$ & MEt3R $\downarrow$ \\
        \midrule
        \rowcolor{gray!10}
        \multicolumn{10}{c}{\textbf{RealEstate10K~\cite{re10k}}} \\
        \midrule
        MVGenMaster~\cite{cao2025mvgenmaster} &   & 15.226 & 0.588 & 0.456 & {0.282} & \textbf{6.42} & \underline{0.526} & \underline{0.664} & {0.339} \\
        Matrix3D~\cite{lu2025matrix3d} &   & 14.490 & 0.580 & \underline{0.448} & 0.413 & 8.93 & 0.638 & 0.666 & 0.344 \\
        CAMEO~\cite{kwon2025cameo} & \ding{55} & 13.800 & 0.561 & 0.522 & 0.446 & 12.93 & 0.790 & \textbf{0.661} & 0.344 \\
        NVComposer~\cite{nvcomposer} &   & 11.140 & 0.418 & 0.649 & 0.829 & 42.85 & 1.435 & 0.860 & 0.457 \\
        CAT3D$^\dagger$~\cite{cat3d} &   & 13.350 & 0.527 & 0.561 & 0.496 & 17.49 & 0.941 & 0.719 & 0.361 \\
        \midrule
        DINO~\cite{dinov2} &   & 15.638 & {0.601} & \underline{0.448} & 0.345 & 15.59 & 0.719 & 0.721 & \textbf{0.319} \\
        VAE~\cite{rombach2022high} & \ding{51} & \underline{15.656} & \underline{0.606} & 0.456 & \underline{0.278} & 8.68 & 0.552 & 0.681 & 0.375 \\
        \textbf{GLD (Ours)} &   & \textbf{16.362} & \textbf{0.630} & \textbf{0.431} & \textbf{0.211} & \underline{7.07} & \textbf{0.444} & {0.673} & \underline{0.328} \\
        \midrule
        \rowcolor{gray!10}
        \multicolumn{10}{c}{\textbf{DL3DV~\cite{ling2024dl3dv}}} \\
        \midrule
        MVGenMaster~\cite{cao2025mvgenmaster} &   & {14.565} & 0.442 & 0.460 & \underline{0.281} & \underline{6.93} & \underline{0.592} & \underline{0.637} & \textbf{0.375} \\
        Matrix3D~\cite{lu2025matrix3d} &   & 13.330 & 0.396 & \underline{0.451} & 0.459 & 9.65 & 0.850 & 0.667 & 0.394 \\
        CAMEO~\cite{kwon2025cameo} & \ding{55} & 12.320 & 0.371 & 0.567 & 1.143 & 24.76 & 2.149 & 0.706 & 0.404 \\
        NVComposer~\cite{nvcomposer} &   & 10.518 & 0.273 & 0.646 & 1.810 & 55.59 & 3.098 & 0.852 & 0.517 \\
        CAT3D$^\dagger$~\cite{cat3d} &   & 11.820 & 0.335 & 0.594 & 1.346 & 31.73 & 2.473 & 0.746 & 0.435 \\
        \midrule
        DINO~\cite{dinov2} &   & 14.345 & {0.411} & 0.471 & 0.546 & 13.12 & 1.050 & 0.708 & 0.410 \\
        VAE~\cite{rombach2022high} & \ding{51} & \underline{14.725} & \underline{0.446} & 0.476 & 0.589 & 15.00 & 1.116 & 0.674 & 0.407 \\
        \textbf{GLD (Ours)} &   & \textbf{15.499} & \textbf{0.468} & \textbf{0.438} & \textbf{0.209} & \textbf{5.75} & \textbf{0.466} & \textbf{0.612} & \underline{0.378} \\
        \midrule
        \rowcolor{gray!10}
        \multicolumn{10}{c}{\textbf{Mip-NeRF 360~\cite{mipnerf} (Out-of-domain)}} \\
        \midrule
        MVGenMaster~\cite{cao2025mvgenmaster} &   & \underline{14.170} & \textbf{0.304} & 0.511 & \textbf{0.320} & \textbf{10.92} & \textbf{0.587} & 0.676 & \underline{0.402} \\
        Matrix3D~\cite{lu2025matrix3d} &   & 13.970 & 0.284 & \textbf{0.483} & \underline{0.548} & \underline{13.63} & \underline{0.948} & \underline{0.646} & 0.422 \\
        CAMEO~\cite{kwon2025cameo} & \ding{55} & 11.900 & 0.250 & 0.629 & 1.623 & 48.75 & 3.008 & 0.684 & \textbf{0.395} \\
        NVComposer~\cite{nvcomposer} &   & 12.525 & 0.217 & 0.637 & 1.622 & 54.26 & 2.703 & 0.767 & 0.526 \\        
        CAT3D$^\dagger$~\cite{cat3d} &   & 11.310 & 0.214 & 0.653 & 1.722 & 54.65 & 3.171 & 0.724 & 0.453 \\
        \midrule
        DINO~\cite{dinov2} &   & 13.718 & 0.267 & 0.542 & 0.949 & 27.57 & 1.720 & 0.707 & 0.444 \\
        VAE~\cite{rombach2022high} & \ding{51} & 13.942 & 0.274 & 0.548 & 1.221 & 35.34 & 2.200 & 0.674 & 0.449 \\
        \textbf{GLD (Ours)} &   & \textbf{14.542} & \underline{0.288} & \underline{0.504} & 0.589 & 15.97 & 1.071 & \textbf{0.630} & {0.406} \\
        \bottomrule
        \end{tabular}%
    }
    \vspace{-10pt}
\end{table*}

%% file: fig/main_qual.tex
\begin{figure}[t!]
  \centering
  \includegraphics[width=\linewidth]{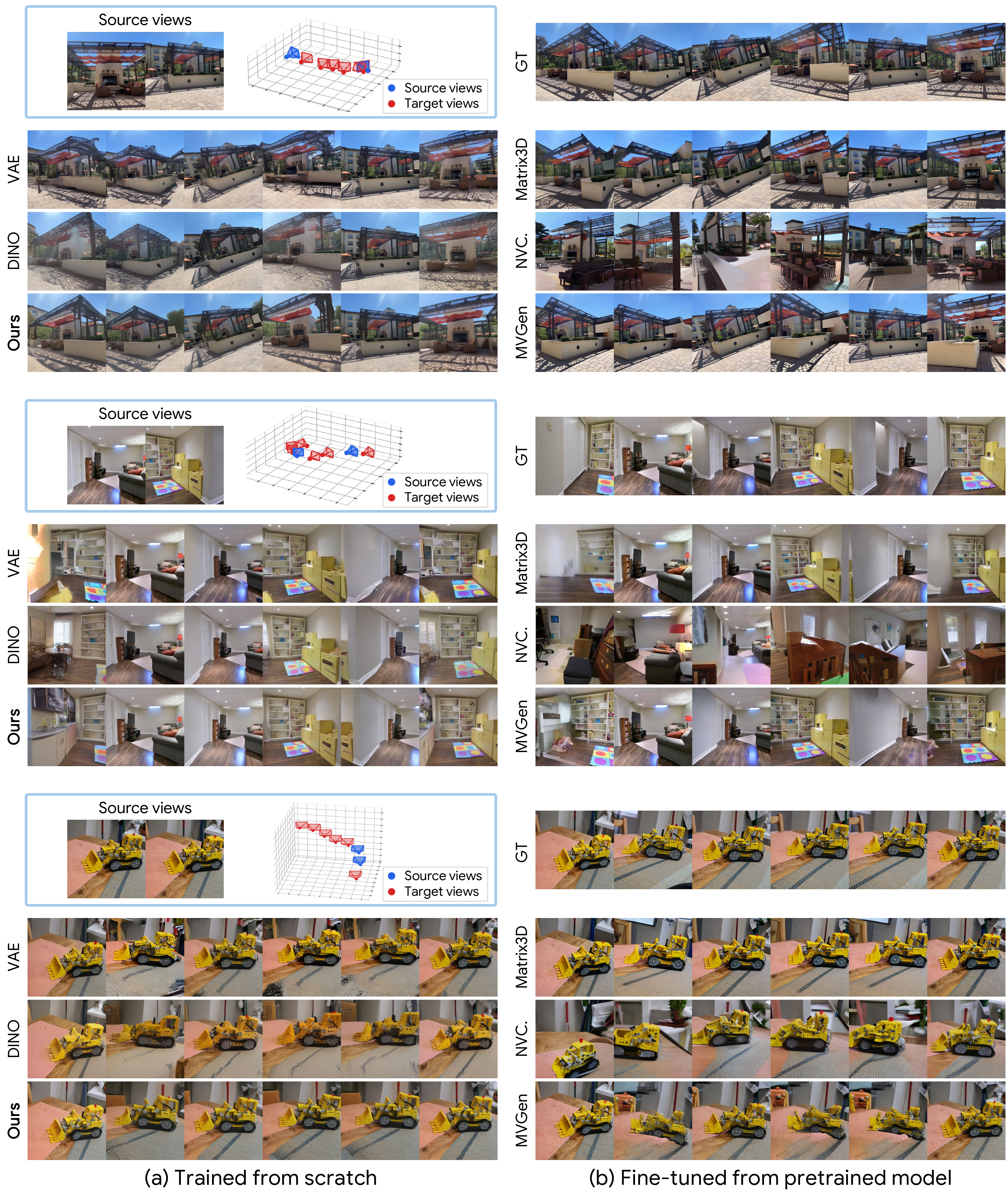} 
  \vspace{-20pt}
  \caption{\textbf{Qualitative results.} We compare the rendering quality of target views given two source views. The rows from top to bottom correspond to the Re10K, DL3DV, and out-of-domain Mip-NeRF 360 datasets, respectively.}
  \label{fig:main_qual}
  \vspace{-20pt}
\end{figure}

%% file: table/v_ablation.tex
\begin{table*}[t]
\centering
\caption{\textbf{Ablation study across different numbers of source images.} We report results with $N{=}1$ and $N{=}4$ input source views on Re10K and DL3DV. \textbf{Bold} values indicate best results.}
\label{tab:cond_ablation}
\vspace{-5pt}
\setlength{\tabcolsep}{6pt}
\small
\renewcommand{\arraystretch}{0.95}
    \resizebox{\textwidth}{!}{%
        \begin{tabular}{c|c|l|ccc|ccccc}
        \toprule
        \multirow{2}{*}{\raisebox{-0.6\height}{\shortstack{Source\\Views}}} & \multirow{2}{*}{\raisebox{-0.44\height}{Dataset}} & \multirow{2}{*}{\raisebox{-0.44\height}{Methods}} & \multicolumn{3}{c|}{2D Metrics} & \multicolumn{5}{c}{3D Metrics} \\
        \cmidrule(lr){4-6} \cmidrule(lr){7-11}
         & & & PSNR $\uparrow$ & SSIM $\uparrow$ & LPIPS $\downarrow$ & ATE $\downarrow$ & RPE$_r$ $\downarrow$ & RPE$_t$ $\downarrow$ & Reproj $\downarrow$ & MEt3R $\downarrow$ \\
        \midrule
        % V=1, Re10K
        & & DINO~\cite{dinov2}  & 13.49 & 0.543 & \textbf{0.541} & 0.442 & 22.95 & 0.972 & 0.735 & \textbf{0.301} \\
        & \multirow{-1}{*}{Re10K} & VAE~\cite{rombach2022high}   & 12.99 & 0.519 & 0.566 & 0.371 & 12.47 & 0.718 & 0.689 & 0.387 \\
        & & \textbf{GLD (Ours)}  & \textbf{13.50} & \textbf{0.552} & \textbf{0.541} & \textbf{0.267} & \textbf{8.42} & \textbf{0.539} & \textbf{0.673} & 0.306 \\
        \cmidrule(l){2-11}
        % V=1, DL3DV
        & & DINO~\cite{dinov2}  & 12.80 & 0.368 & \textbf{0.539} & 0.743 & 20.14 & 1.405 & 0.738 & 0.418 \\
        \multirow{-5}{*}{$N{=}1$} & \multirow{-1}{*}{DL3DV} & VAE~\cite{rombach2022high}   & 12.47 & 0.360 & 0.569 & 0.880 & 25.93 & 1.828 & 0.702 & 0.420 \\
        & & \textbf{GLD (Ours)}  & \textbf{13.03} & \textbf{0.385} & 0.543 & \textbf{0.237} & \textbf{6.14} & \textbf{0.512} & \textbf{0.619} & \textbf{0.375} \\
        \midrule
        % V=4, Re10K
        & & DINO~\cite{dinov2}  & 17.73 & 0.651 & 0.365 & 0.221 & 7.92 & 0.482 & 0.685 & \textbf{0.327} \\
        & \multirow{-1}{*}{Re10K} & VAE~\cite{rombach2022high}   & 18.68 & 0.695 & 0.348 & 0.200 & \textbf{6.26} & 0.407 & 0.655 & 0.347 \\
        & & \textbf{GLD (Ours)}  & \textbf{19.00} & \textbf{0.696} & \textbf{0.327} & \textbf{0.182} & 6.69 & \textbf{0.397} & \textbf{0.653} & \textbf{0.327} \\
        \cmidrule(l){2-11}
        % V=4, DL3DV
        & & DINO~\cite{dinov2}  & 15.44 & 0.443 & 0.423 & 0.274 & 7.49 & 0.563 & 0.667 & 0.408 \\
        \multirow{-5}{*}{$N{=}4$} & \multirow{-1}{*}{DL3DV} & VAE~\cite{rombach2022high}   & 16.37 & 0.505 & 0.410 & 0.294 & 8.14 & 0.556 & 0.643 & 0.406 \\
        & & \textbf{GLD (Ours)}  & \textbf{17.09} & \textbf{0.517} & \textbf{0.378} & \textbf{0.143} & \textbf{3.74} & \textbf{0.305} & \textbf{0.601} & \textbf{0.385} \\
        \bottomrule
        \end{tabular}%
    }
    \vspace{-10pt}
\end{table*}

%% file: fig/level_analysis.tex
\begin{figure}[t]
    \centering
    
    \begin{minipage}[t]{0.48\textwidth}
        \centering
        \vspace{-5pt}
        % --- TOP TABLE (transposed) ---
        \captionof{table}{\textbf{Geometric Correspondence.} PCK results for each DA3 level and DINOv2~\cite{dinov2}.}
        \label{tab:pck_eval}
        {\setlength{\tabcolsep}{6pt}
        \resizebox{\linewidth}{!}{%
        \begin{tabular}{@{}l|ccccc@{}}
            \toprule
             Feature & $F_0$ & $F_1$ & $F_2$ & $F_3$ & DINOv2 \\
            \midrule
            PCK $\uparrow$ & 22.25 & \underline{35.98} & \textbf{40.70} & 20.98 & 31.64 \\
            \bottomrule
        \end{tabular}%
        }}
        
        % \vspace{1em}
        
       % --- BOTTOM TABLE (transposed) ---
        \captionof{table}{\textbf{Reconstruction Fidelity.} Reconstruction comparison via PSNR and LPIPS across different DA3~\cite{da3} levels.}
        \label{tab:recon_eval}
        \scalebox{0.7}{%
        {\setlength{\tabcolsep}{10pt}
        \begin{tabular}{@{}l|cccc@{}}
            \toprule
            Feature & $F_0$ & $F_1$ & $F_2$ & $F_3$ \\
            \midrule
            PSNR $\uparrow$ & \textbf{28.01} & \underline{25.36} & 14.01 & 10.19 \\
            SSIM $\uparrow$ & \textbf{0.922} & \underline{0.873} & 0.627 & 0.508 \\
            LPIPS $\downarrow$ & \textbf{0.138} & \textbf{0.138} & \underline{0.491} & 0.768 \\
            \bottomrule
        \end{tabular}
        }}
    \end{minipage}\hfill
    \begin{minipage}[t]{0.48\textwidth}
        \vspace{0pt}
        \vspace*{5pt}
        \centering
        \includegraphics[width=\linewidth]{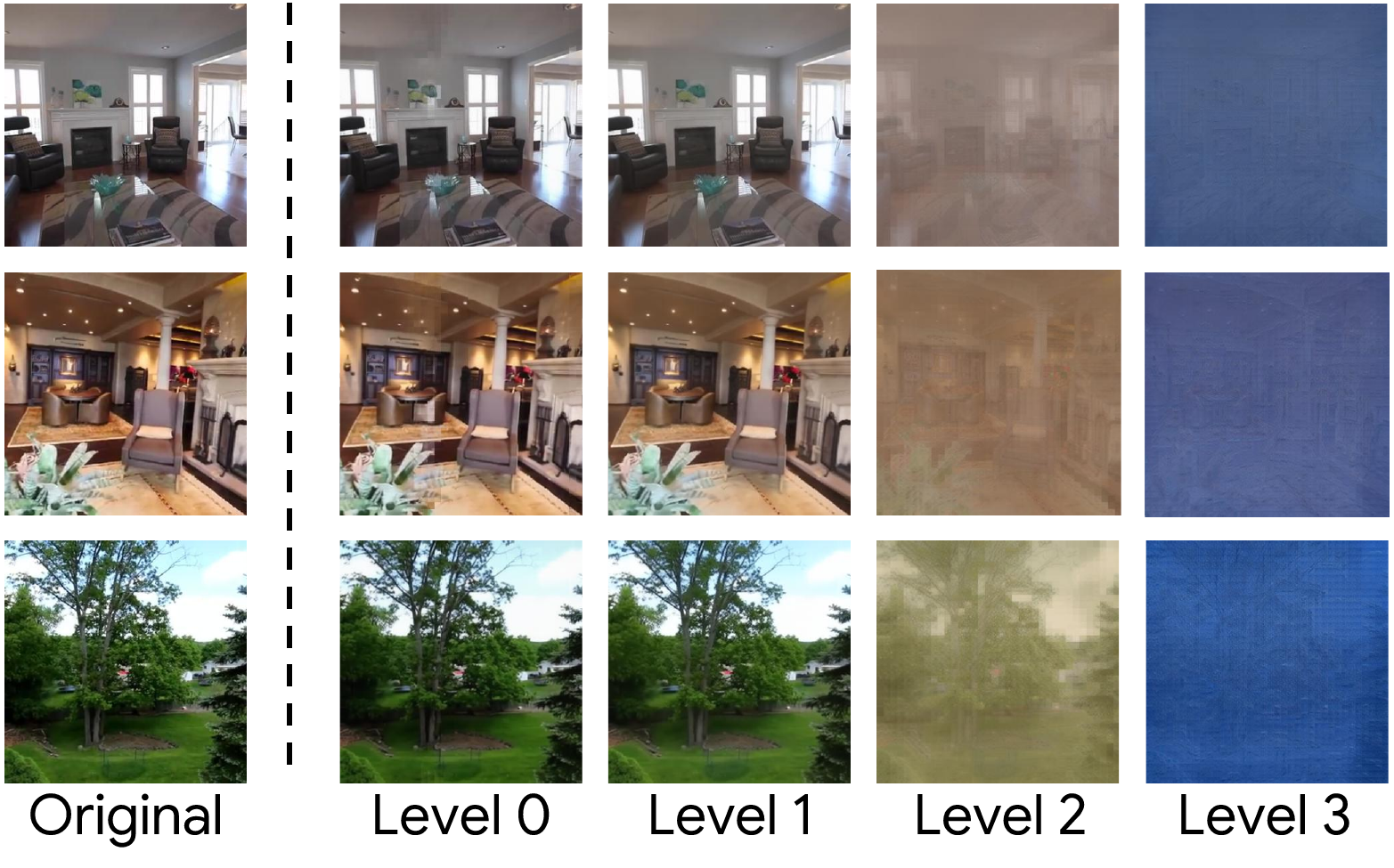}
        \setlength{\abovecaptionskip}{-5pt}
        \captionof{figure}{\textbf{Reconstruction Visualization.} Qualitative comparison showing the reconstruction results for different levels.}
        \label{fig:recon_vis}
    \end{minipage}
    \vspace{-15pt}
\end{figure}

%% file: table/reconstruction.tex
\begin{wraptable}[8]{r}{0.5\textwidth}
  \centering
  \vspace{-33pt}
\caption{\textbf{Reconstruction Fidelity Comparison.} We evaluate the reconstruction quality using 4,000 randomly sampled images from the Re10K~\cite{re10k} test set.}
  \label{tab:recon_comparison}
  % \linewidth here refers to the width of the wraptable container (0.5\textwidth)
  \setlength{\tabcolsep}{14pt}
  \resizebox{1.0\linewidth}{!}{%
    \begin{tabular}{l|ccc}
      \toprule
      Methods & PSNR $\uparrow$& SSIM $\uparrow$ & LPIPS $\downarrow$ \\
      \midrule
      VAE~\cite{rombach2022high}       & 34.53 &0.939 &0.028 \\
      VAE  (SDXL)~\cite{podell2023sdxl}          & 34.97& 0.945& 0.029  \\
      RAE (DINO)~\cite{rae}       & 26.78 & 0.830 &0.148  \\
      \midrule
      \textbf{DA3 Dec. ($\mathcal{D}_\text{rgb}$)}        & \textbf{35.41}& \textbf{0.960} & \textbf{0.019} \\
      \bottomrule
    \end{tabular}%
  }
  \vspace{-20pt}
\end{wraptable}

%% file: table/independent.tex
% \begin{wraptable}{r}{0.48\textwidth}
%   \centering
%   \vspace{-35pt}
%   \caption{\textbf{Cascading vs Independent Generation.} We evaluate the reconstruction quality using 4,000 randomly sampled images from the RealEstate10K test set.}
%   \label{tab:cascading}
%   % \linewidth here refers to the width of the wraptable container (0.5\textwidth)
%   \setlength{\tabcolsep}{8pt}
%   \resizebox{1.0\linewidth}{!}{%
%     \begin{tabular}{l|ccc}
%       \toprule
%       \textbf{Method} & PSNR $\uparrow$& SSIM $\uparrow$ & LPIPS $\downarrow$ \\
%       \midrule
%       Independent       & 18.8135&  0.6925& 0.3357\\
%       Cascading (Ours)        &  19.0029& 0.6958&0.3272\\
%       \bottomrule
%     \end{tabular}%
%   }
%   \vspace{-20pt}
%   \label{tab:reconstruction}
% \end{wraptable}

\begin{table}[t]
  \centering
  \caption{\textbf{Cascaded vs. Independent Generation.} We evaluate NVS performance on Re10K with $N{=}4$ source views.}
  \vspace{-10pt}
  \label{tab:cascading}
  \setlength{\tabcolsep}{6pt}
  \resizebox{\textwidth}{!}{%
    \begin{tabular}{l|ccc|ccccc}
      \toprule
      & \multicolumn{3}{c|}{2D Metrics} & \multicolumn{5}{c}{3D Metrics} \\
      Methods & PSNR $\uparrow$ & SSIM $\uparrow$ & LPIPS $\downarrow$ & ATE $\downarrow$ & RPE$_r$ $\downarrow$ & RPE$_t$ $\downarrow$ & Reproj $\downarrow$ & MEt3R $\downarrow$ \\
      \midrule
      Independent    & 18.81 & 0.692 & 0.335 & 0.197 & 7.179 & 0.430 & 0.666 & 0.335 \\
      Cascaded (Ours) & \textbf{19.00} & \textbf{0.695} & \textbf{0.327} & \textbf{0.182} & \textbf{6.694} & \textbf{0.397} & \textbf{0.652} & \textbf{0.326} \\
      \bottomrule
    \end{tabular}%
  }
  \vspace{-18pt}
\end{table}

%% file: table/depth.tex
\begin{wraptable}[7]{r}{0.5\linewidth}               
\vspace{-33pt}
\centering   
    \caption{\textbf{Depth Evaluation.} Comparison with Matrix3D~\cite{lu2025matrix3d} on ETH3D~\cite{eth}.}
    \vspace{5pt}
    \label{tab:eth3d_cond2}                          
  \resizebox{\linewidth}{!}{%
  \setlength{\tabcolsep}{3pt}
  \begin{tabular}{l|ccc|c}
  \toprule
  \multirow{2}{*}{Methods}
    & \multicolumn{3}{c|}{Depth}
    & RGB \\
    & AbsRel$\downarrow$ & SqRel$\downarrow$ & $\delta_1$$\uparrow$
    & PSNR$\uparrow$ \\
  \midrule
  Matrix3D~\cite{lu2025matrix3d}  & 0.197 & 0.475 & 0.731 & 14.13 \\
  \textbf{GLD (Ours)}         & \textbf{0.160} & \textbf{0.410} & \textbf{0.800} & \textbf{14.80} \\
  \bottomrule
  \end{tabular}}
    % \vspace{15pt}
  \end{wraptable}

%% file: fig/3d_vis.tex
\begin{figure}[t]
  \centering
  \setlength{\tabcolsep}{2pt} % 좌우 간격(필요시 조절)
  \begin{tabular}{cc}
    \includegraphics[width=0.49\linewidth]{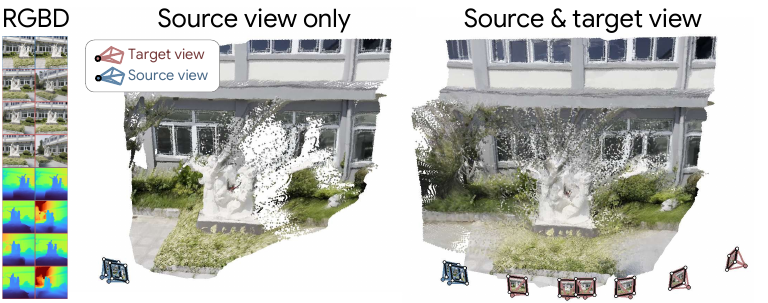} &
    \includegraphics[width=0.49\linewidth]{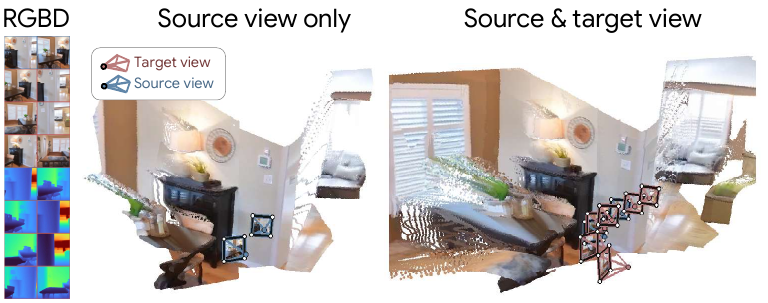} \\
    \includegraphics[width=0.49\linewidth]{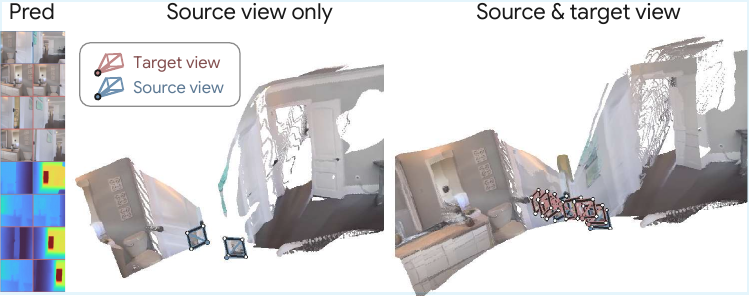} &
    \includegraphics[width=0.49\linewidth]{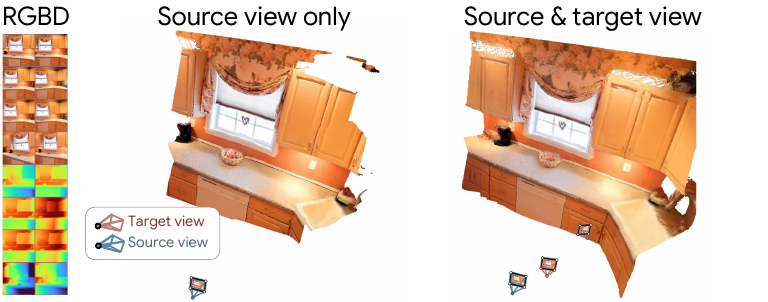}
  \end{tabular}
  \vspace{-5pt} % 필요하면 조절
  \caption{\textbf{3D Reconstruction Visualization.} We unproject the synthesized novel views into 3D space using the jointly generated depth and ray maps.}
  \label{fig:reconstruction_vis}
  \vspace{-5pt}
\end{figure}

\begin{figure}[t]
  \centering
  \includegraphics[width=1.0\linewidth]{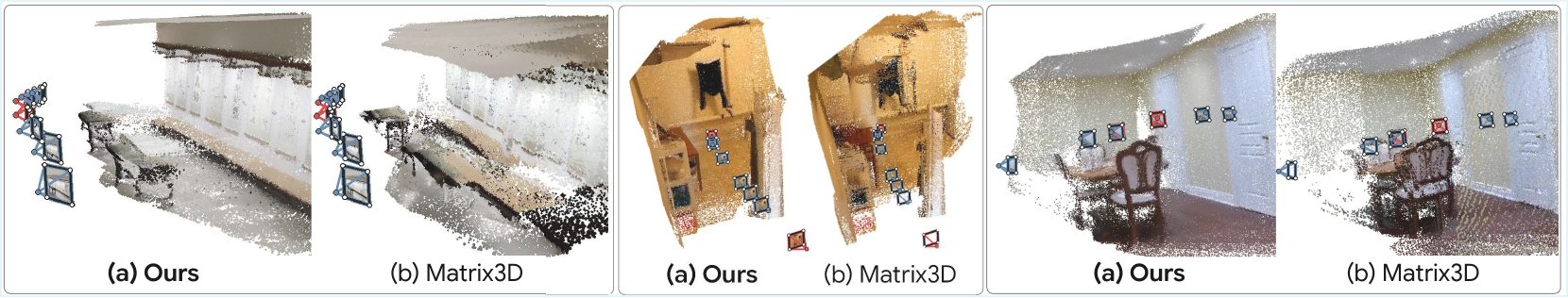}
  %\vspace{-10pt} % 필요하면 조절
  \caption{\textbf{3D Reconstruction Comparison:} Point clouds unprojected from source and target views RGBD outputs of GLD and Matrix3D~\cite{lu2025matrix3d}.}
  \label{fig:3d_compare}
  \vspace{-15pt}
\end{figure}

%% file: sec/7_conclusion.tex
%\vspace{-15pt}
\section{Conclusion}
\label{sec:conclusion}
We presented \textbf{Geometric Latent Diffusion (GLD)}, a framework that repurposes the feature space of a geometric foundation model as the latent space for novel view synthesis. Through systematic analysis, we identified a feature level that balances geometric correspondence and photometric fidelity and showed that training diffusion models in this space yields consistent improvements in both 2D image quality and 3D consistency over standard VAE and DINOv2 latent spaces. GLD achieves competitive performance with state-of-the-art methods that leverage large-scale text-to-image
pretraining, despite being trained entirely from scratch, while also enabling zero-shot depth and 3D reconstruction as a natural
byproduct of the generation process. We hope that this work encourages further investigation into task-specific latent space design for geometry-aware generation.

\section*{Acknowledgment}
This work was completed during WJ and JH’s visit to NYU as part of the Global AI Frontier Lab Program.
S.X. acknowledges support from Intel Labs, the MSIT IITP grant (RS-2024-00457882) and the NSF award IIS-2443404.

%% file: sec/appendix.tex
% \textbf{\textcolor{red}{NOTE!! This should be separated and removed in the main pdf before submission.}}
% ---------------------------------------------------------------
% Appendix — starts a new section counter (A, B, C, …)
% ---------------------------------------------------------------

\setcounter{section}{0}
\renewcommand{\thesection}{\Alph{section}}

\section*{\Large Appendix}

% =================================================================
%  A.  Implementation Details
% =================================================================
\section{Implementation Details}
\label{app:implementation}

\subsection{Training Details}
\label{app:training}

% fixed lr, dropout prob. included
% cam drop (PRoPE and Plucker), dataset ratio, varing image resolution
% learning rate schedule, augmentation, dropout rates for conditioning pathways, etc.

This subsection describes the training configurations. Hyperparameters are summarized in~\cref{tab:decoder_discriminator_config}.

\subsubsection{RGB Decoder.}
We train the RGB decoder $\mathcal{D}_{\text{rgb}}$ on images of varying resolutions, including (504$\times$504), (504$\times$378), (504$\times$336), and (504$\times$280). Following RAE~\cite{rae}, we optimize the decoder using $\ell_1$, LPIPS~\cite{zhang2018unreasonable}, and GAN losses~\cite{sauer2023stylegan}. The reconstruction losses are weighted equally, while the GAN loss is scaled with an adaptive weight~\cite{esser2021taming} to balance reconstruction and adversarial supervision. For the discriminator, we adopt the same setup as StyleGAN-T~\cite{sauer2023stylegan} and apply differentiable augmentations~\cite{zhao2020differentiable} before feeding images to the discriminator.

\subsubsection{Multi-view Diffusion.} For classifier-free guidance (CFG)~\cite{ho2022classifier}, camera embeddings are dropped by zeroing the Plücker ray embeddings and setting the extrinsic matrices to identity in PRoPE~\cite{prope}. To handle varying image resolutions, input resolutions are sampled per batch from (504$\times$504), (504$\times$378), (504$\times$336), and (504$\times$280) with ratio $2:2:1:1$. Each training batch consists of 48 scenes, each containing $V=8$ views.

Some vision encoders produce additional non-spatial tokens, such as register tokens in DINOv2~\cite{dinoregister, dinov2} and camera tokens in VGGT~\cite{wang2025vggt} and DA3~\cite{da3}. While our architecture can jointly denoise these tokens via dedicated embedders, we omitted them from our final experiments for simplicity, as the downstream decoders operate exclusively on spatial features.

\subsection{Architecture Details}
\label{app:architecture}
\input{fig/suppl/full_architecture}
% \ref{sec:select_level}
This subsection details the implementation of each module introduced in \S\textcolor{red}{4}. The full architecture is illustrated in \cref{fig:suppl_full-architecture}.
\subsubsection{RGB Decoder.}
We utilize a ViT-based RGB decoder with a patch size of 14. The model is built with 12 transformer layers, an intermediate dimension of 3072, and a dropout probability of 0.5. The decoder learns to reconstruct RGB images from the geometric latent features. The architecture is illustrated in \cref{fig:suppl_full-architecture}(B).

\subsubsection{Multi-Level Feature Extraction.}
Given $V$ input views, the frozen geometric encoder $\mathcal{E}_{\text{geo}}(\cdot)$ extracts $L{=}4$ feature levels, where $T$ is the token sequence length at the encoder patch resolution and $C{=}1536$ is the channel dimension.
The four levels correspond to intermediate outputs of the frozen DA3-Base backbone after blocks $\{5,7,9,11\}$, matching the multi-scale outputs consumed by the DA3 decoder.
Before diffusion training, each level is normalized to zero mean and unit variance using channel-wise statistics precomputed on the training set. The inverse transform is applied before decoding.

\subsubsection{Multi-view Diffusion Model.}
As illustrated in \cref{fig:suppl_full-architecture}(A), each level-wise diffusion model follows the $\text{DiT}^\text{DH}$~\cite{ddt} architecture, which decouples the network into two components: a \textbf{condition encoder} and a \textbf{velocity decoder}.

The condition encoder processes the input tokens with a hidden dimension $C_1{=}768$ through $28$ DiT blocks, producing a compact representation that summarizes the source--target context.
The velocity decoder uses a larger hidden dimension $C_2{=}2048$ and predicts the velocity field $\mathbf{u}_{t,l} \in \mathbb{R}^{V \times T \times C}$.
For the level-wise models $\mathcal{M}_l$, the velocity decoder consists of $6$ DiT blocks, whereas for the cascaded model $\mathcal{M}_{1 \to 0}$, it consists of $2$ DiT blocks.
The encoder's output conditions the decoder via AdaLN modulation, and camera information is injected into the decoder through addition.
All models share the same overall design, while the cascaded model $\mathcal{M}_{1 \to 0}$ uses a shallower velocity decoder.

\begin{table}[t]
\centering
\caption{\textbf{DiT$^\text{DH}$ configuration used for all level-wise diffusion models in GLD.}}
\label{tab:appendix_dit}
\small
\setlength{\tabcolsep}{10pt}
\begin{tabular}{lcc}
\toprule
\textbf{Parameter} & \textbf{Condition Encoder} & \textbf{Velocity Decoder} \\
\midrule
Hidden dimension   & $C_1 = 768$  & $C_2 = 2048$ \\
Number of blocks   & 28   & 6 ($\mathcal{M}_l$), 2 ($\mathcal{M}_{1 \to 0}$)    \\
Attention heads    & 16   & 16   \\
MLP activation     & SwiGLU & SwiGLU \\
Normalization      & RMSNorm & RMSNorm \\
\midrule
Input channels     & \multicolumn{2}{c}{ $2C=3072$ ($\mathcal{M}_l$)} \\
Output channels    & \multicolumn{2}{c}{$C = 1536$} \\
Patch size & \multicolumn{2}{c}{1 (DA3, DINO), 2 (VAE)} \\
Positional encoding & \multicolumn{2}{c}{PRoPE~\cite{prope}} \\
\bottomrule
\end{tabular}
\end{table}

The model receives the noisy latent $\mathbf{z}_{t}^{l}$ concatenated channel-wise with the source-only condition $\mathbf{F}^{\text{src}}_l$ (zero-padded for target views). For the cascaded model $\mathcal{M}_{1 \to 0}$, the condition encoder receives the source-only condition and the deeper level features $\mathbf{F}_{l+1}$. Since source views have clean features while target views are zero-padded, we use separate patch embedders for each view type in both the encoder and decoder, yielding four embedders in total. These embedders tokenize the latent feature maps and project them from $2C$ channels to $C_1$ or $C_2$: for DA3 latents, we use a patch size of 1 to preserve the original token resolution, whereas for VAE latents, we use a patch size of 2 to reduce the token count due to their higher spatial resolution.

After embedding, all view tokens are concatenated along the view dimension and processed jointly through DiT blocks~\cite{ddt}, in which standard self-attention is replaced with 3D self-attention to enable cross-view interactions.

For camera conditioning, we compute per-pixel 6D Pl\"ucker coordinates from the camera intrinsics and extrinsics, concatenate a binary source/target indicator $\mathbf{m} \in \{0,1\}^{V \times T \times 1}$ (0 for source views and 1 for target views) to form a 7D embedding, and project it to the hidden dimension via a linear layer. PRoPE~\cite{prope} is applied in every 3D self-attention layer.

\begin{table}[h]
      \centering
      \caption{\textbf{Training configuration for decoder and discriminator.}}
      \label{tab:decoder_discriminator_config}
      \setlength{\tabcolsep}{10pt}
      \begin{tabular}{lll}
          \toprule
          \textbf{Component} & \textbf{Decoder} $\mathcal{D_\text{rgb}}$ & \textbf{Multi-view Diffusion} \\
          \midrule
          optimizer                & Adam                   & AdamW                 \\
          max learning rate        & $2 \times 10^{-4}$     & $5 \times 10^{-5}$    \\
          min learning rate        & $2 \times 10^{-5}$     & $5 \times 10^{-5}$    \\
          learning rate schedule   & cosine decay           & constant              \\
          optimizer betas          & $(0.9,\,0.95)$          & $(0.9,\,0.95)$        \\
          weight decay             & 0.0                    & 0.0                   \\
          batch size               & 16                    & 48                    \\
          warmup                   & 1 epoch                & --                    \\
          loss                     & $\ell_1$ + LPIPS +  GAN & v-prediction          \\
          Model                    & ViT-XL        & DiT$^\text{DH}$  \\
          EMA decay                & 0.9978                     & 0.9995                \\
          gradient clipping        & --                     & 1.0                   \\
          \bottomrule
      \end{tabular}
  \end{table}

% additional details on model architeuctues and implemnentation (ProPE, layer num, ...)
% should include Plucker ray, omitting text embedder, ...

\subsection{Dataset Details}
\label{app:dataset}

\subsubsection{RGB Decoder.} We train the RGB decoder on Re10K~\cite{re10k} and DL3DV~\cite{ling2024dl3dv}, sampling the two datasets at an equal ratio. For evaluation, we randomly sample 500 scenes, each with 8 views, resulting in 4,000 images in total. All images are resized to a resolution of $504 \times 504$.

\begin{table}[t]
\centering
\caption{\textbf{Number of scenes per dataset used for training and evaluation.}}
\label{tab:dataset}
    \setlength{\tabcolsep}{10pt}
    \begin{tabular}{lcc}
        \toprule
        Dataset & Train & Eval \\
        \midrule
        Re10K~\cite{re10k} & 66,033 & 200 \\
        DL3DV~\cite{ling2024dl3dv}         & 10,176 & 55 \\
        HyperSim~\cite{hypersim}      & 794 & --  \\
        TartanAir~\cite{tartanair2020iros}     & 369 & --  \\
        MipNeRF-360~\cite{mipnerf}   & -- & 9 \\
        ETH3D~\cite{eth}         & -- & 13 \\
        \bottomrule
    \end{tabular}
\end{table}

% dataset size, view setups for each dataset for train/inference
\subsubsection{Multi-view Diffusion Model.} We train GLD on four datasets: Re10K~\cite{re10k}, DL3DV~\cite{ling2024dl3dv}, HyperSim~\cite{hypersim}, and TartanAir~\cite{tartanair2020iros}, with a mixing ratio of $4:4:1:1$. We use Mip-NeRF 360~\cite{mipnerf} for out-of-domain evaluation and ETH3D~\cite{eth} for depth evaluation. To handle varying scene scales across datasets, all camera poses are normalized relative to the last view, which is set as the origin, and scaled such that the maximum camera distance within the batch is 1. The multi-view sequence starts from a randomly sampled frame index chosen to ensure sufficient remaining frames for the required number of views, with each consecutive frame interval independently and uniformly sampled within a predefined range. For interpolation, the first and last frames are always included as source views when $N=2$, while additional source views are placed at uniform intervals when $N>2$.

For evaluation, we use 200 samples per dataset from Re10K, DL3DV, and Mip-NeRF 360 for NVS, and 50 samples from ETH3D for depth evaluation. For datasets whose evaluation sets contain fewer scenes than the target count~\cite{ling2024dl3dv,eth,mipnerf}, multiple samples are drawn from the same scene using randomized view sampling. Dataset statistics are summarized in~\cref{tab:dataset}.

% =================================================================
%  B.  Evaluation Details
% =================================================================
\section{Evaluation Details}
\label{app:evaluation}

\subsection{Baseline Adaptation}
\label{app:baselines}
In this section, we describe how each baseline is adapted for our evaluation. For all baselines, we use the default CFG scale provided in the official repositories. We also resize the generated images from $512 \times 512$ to $504 \times 504$ to match GLD's output resolution and ensure a fair comparison.
\subsubsection{MVGenMaster.} MVGenMaster~\cite{cao2025mvgenmaster} requires metric depth for the reference views, warped RGB, and warped depth for the target view as input conditioning. Since metric depth is not provided in their evaluation setting, we leverage Depth Anything 3~\cite{da3} to estimate metric depth for the reference view, aligned with the ground-truth extrinsics. We then warp the reference RGB and depth into the target view using the GT camera pose, yielding the final input conditionings for the model. 

\subsubsection{CAT3D$^\dagger$.}
% Both CAMEO~\cite{kwon2025cameo} and CAT3D$^\dagger$ require relative camera poses as conditioning, which we obtain from VGGT~\cite{wang2025vggt}-predicted poses. 
As the official implementation of CAT3D is not publicly available, we denote by CAT3D$^\dagger$ a reproduction using the model architecture and checkpoint from CAMEO~\cite{kwon2025cameo}.

% \subsubsection{Matrix3D.} Following the original paper, we set the CFG scale of Matrix3D~\cite{lu2025matrix3d} to 1.5, and interpolate the outputs from $512{\times}512$ to $504{\times}504$.

% \subsubsection{NVComposer.} Following the official implementation, we set the CFG scale of NVComposer~\cite{nvcomposer} to 7.5, and interpolate the outputs from $512{\times}512$ to $504{\times}504$.

% \subsection{Benchmark Construction}
% \label{app:benchmark}
% eval view setup 

% =================================================================
%  C.  Additional Results
% =================================================================
\begin{figure}[t]
    \centering
    \includegraphics[width=\textwidth]{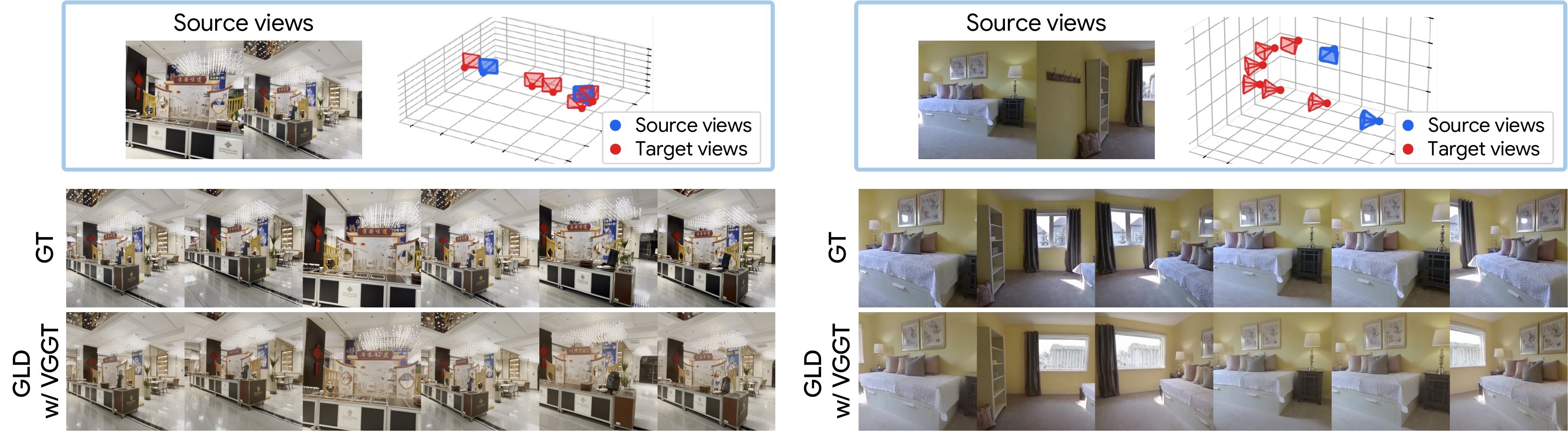}
    \vspace{-15pt}
    \caption{\textbf{Qualitative results of GLD w/ VGGT.}}
    \label{supfig:vggt_qual}
    \vspace{-5pt}
\end{figure}

\begin{table*}[t]
\centering
\caption{Quantitative comparison of GLD using DA3~\cite{da3} and VGGT~\cite{wang2025vggt} backbones against baseline latent representations across in-domain~\cite{re10k, ling2024dl3dv} and out-of-domain~\cite{mipnerf} datasets on 2D and 3D metrics. \textbf{Bold} and \underline{underlined} values indicate the best and second-best results, respectively.}
\label{supptab:vggt}
\vspace{-5pt}
\setlength{\tabcolsep}{7pt}
\small
\renewcommand{\arraystretch}{0.85}
    \resizebox{\textwidth}{!}{%
        \begin{tabular}{l|ccc|ccccc}
        \toprule
        \multirow{2}{*}{Methods} & \multicolumn{3}{c|}{2D Metrics} & \multicolumn{5}{c}{3D Metrics} \\
        \cmidrule(lr){2-4} \cmidrule(lr){5-9}
         & PSNR $\uparrow$ & SSIM $\uparrow$ & LPIPS $\downarrow$ & ATE $\downarrow$ & RPE$_r$ $\downarrow$ & RPE$_t$ $\downarrow$ & Reproj $\downarrow$ & MEt3R $\downarrow$ \\
        \midrule
        \rowcolor{gray!10}
        \multicolumn{9}{c}{RealEstate10K~\cite{re10k}} \\
        \midrule
        DINO~\cite{dinov2} & 15.64 & 0.601 & 0.448 & 0.345 & 15.59 & 0.719 & 0.721 & \textbf{0.319} \\
        VAE~\cite{rombach2022high} & 15.66 & \underline{0.606} & 0.456 & 0.278 & 8.68 & 0.552 & 0.681 & 0.375 \\
        \rowcolor{lightpurblue}
        GLD w/ VGGT~\cite{wang2025vggt} & \underline{16.17} & 0.596 & \textbf{0.429} & \underline{0.216} & \underline{7.17} & \textbf{0.440} & \textbf{0.666} & \underline{0.325} \\
        \rowcolor{lightpurblue}
        GLD w/ DA3~\cite{da3} & \textbf{16.36} & \textbf{0.630} & \underline{0.431} & \textbf{0.211} & \textbf{7.07} & \underline{0.444} & \underline{0.673} & 0.328 \\
        \midrule
        \multicolumn{9}{c}{DL3DV~\cite{ling2024dl3dv}} \\
        \midrule
        DINO~\cite{dinov2} & 14.35 & 0.411 & 0.471 & 0.546 & 13.12 & 1.050 & 0.708 & 0.410 \\
        VAE~\cite{rombach2022high} & 14.73 & \underline{0.446} & 0.476 & 0.589 & 15.00 & 1.116 & 0.674 & 0.407 \\
        \rowcolor{lightpurblue}
        GLD w/ VGGT~\cite{wang2025vggt} & \underline{15.25} & 0.434 & \textbf{0.436} & \textbf{0.188} & \textbf{5.23} & \textbf{0.462} & \underline{0.634} & \underline{0.386} \\
        \rowcolor{lightpurblue}
        GLD w/ DA3~\cite{da3} & \textbf{15.50} & \textbf{0.468} & \underline{0.438} & \underline{0.209} & \underline{5.75} & \underline{0.466} & \textbf{0.612} &  \textbf{0.378}\\
        \midrule
        \rowcolor{gray!10}
        \multicolumn{9}{c}{Mip-NeRF 360~\cite{mipnerf} (Out-of-domain)} \\
        \midrule
        DINO~\cite{dinov2} & 13.72 & 0.267 & 0.542 & 0.949 & 27.57 & 1.720 & 0.707 & 0.444 \\
        VAE~\cite{rombach2022high} & \underline{13.94} & \underline{0.274} & 0.548 & 1.221 & 35.34 & 2.200 & 0.674 & 0.449 \\
        \rowcolor{lightpurblue}
        GLD w/ VGGT~\cite{wang2025vggt} & 13.57 & 0.265 & \underline{0.529} & \underline{0.596} & \underline{16.58} & \underline{1.190} &  \underline{0.654} & \textbf{0.394} \\
        \rowcolor{lightpurblue}
        GLD w/ DA3~\cite{da3} & \textbf{14.54} & \textbf{0.288} & \textbf{0.504} & \textbf{0.589} & \textbf{15.97} & \textbf{1.071} &\textbf{0.630} &  \underline{0.405} \\
        \bottomrule
        \end{tabular}%
    }
    \vspace{-10pt}
\end{table*}

\section{Additional Results}
\label{app:results}
\subsection{Evaluation with VGGT Backbone}
\label{app:vggt}

While our primary evaluation established the effectiveness of the Geometric Latent Diffusion (GLD) framework using Depth Anything 3 (DA3)~\cite{da3} as the underlying backbone, we further validate that our core hypothesis generalizes beyond a single specific architecture by evaluating GLD with an alternative geometric foundation model, VGGT~\cite{wang2025vggt}.

Similar to DA3, VGGT extracts feature representations that are deeply grounded in 3D geometry and multi-view consistency. In this experiment, we replace the DA3 encoder and decoder with those of VGGT, repurposing its intermediate feature space as the latent space for our multi-view diffusion model. The diffusion models are trained from scratch following the exact same configuration and objective described in Appendix~\ref{app:implementation}.

As illustrated in \cref{supptab:vggt} and \cref{supfig:vggt_qual}, although GLD with the VGGT backbone exhibits slightly lower overall performance compared to the DA3 backbone, it consistently outperforms the VAE and DINO baselines. This performance gap is particularly pronounced in 3D evaluation metrics, where the VGGT backbone demonstrates strong geometric consistency and significantly lower pose estimation errors (\eg, ATE) across multiple datasets. These findings confirm that operating within a geometric latent space inherently provides the essential multi-view correspondences and 3D structural priors required for robust novel view synthesis, independent of the specific geometric foundation model architecture.

\subsection{Comparison with Method Trained from Scratch}
\label{app:mvgen}
In the main manuscript, we show that our method remains competitive with, and in some cases outperforms, state-of-the-art novel view synthesis (NVS) baselines that leverage large-scale text-to-image (T2I) pretraining. To provide a more controlled and fair comparison, we additionally consider a baseline trained from scratch without any pretrained T2I prior. 

Among the baselines included in our study, MVGenMaster~\cite{cao2025mvgenmaster} is the only method with publicly available training code. We therefore train MVGenMaster from scratch using the same training setup as our method, enabling a controlled comparison that removes the advantage of large-scale generative pretraining.

We compare the methods on the Mip-NeRF 360~\cite{mipnerf} dataset, which lies outside our training domain and remains challenging for GLD. As shown in \cref{supptab:mvgen_scratch}, MVGenMaster trained from scratch performs substantially worse than both its finetuned version and GLD. We attribute this gap to its limited robustness to misalignment between depth and camera parameters, which can induce warping errors during generation. In contrast, the finetuned MVGenMaster benefits from stronger image priors inherited from text-to-image pretraining, making it more tolerant to such errors.

\begin{table*}[t]
\centering
\caption{Quantitative comparison of MVGenMaster~\cite{cao2025mvgenmaster} trained from scratch and GLD (Ours) on out-of-domain~\cite{mipnerf} datasets.}
\label{supptab:mvgen_scratch}
\vspace{-5pt}
\setlength{\tabcolsep}{5pt}
\small
\renewcommand{\arraystretch}{0.85}
    \resizebox{\textwidth}{!}{%
        \begin{tabular}{l|c|ccc|ccccc}
        \toprule
        \multirow{2}{*}{Methods} & \multirow{2}{*}[-2pt]{\shortstack{From\\Scratch}} & \multicolumn{3}{c|}{2D Metrics} & \multicolumn{5}{c}{3D Metrics} \\
        \cmidrule(lr){3-5} \cmidrule(lr){6-10}
         & & PSNR $\uparrow$ & SSIM $\uparrow$ & LPIPS $\downarrow$ & ATE $\downarrow$ & RPE$_r$ $\downarrow$ & RPE$_t$ $\downarrow$ & Reproj $\downarrow$ & MEt3R $\downarrow$ \\
        \midrule
        \rowcolor{gray!10}
        \multicolumn{10}{c}{Mip-NeRF 360~\cite{mipnerf} (Out-of-domain)} \\
        \midrule
        MVGenMaster~\cite{cao2025mvgenmaster} & \ding{55}  & 14.170 & 0.304 & 0.511 & 0.320 & 10.92 & 0.587 & 0.676 & 0.402 \\
        \midrule
        MVGenMaster~\cite{cao2025mvgenmaster} & \ding{51} & 11.217 & 0.197 & 0.665 & 1.804 & 61.291 & 3.423 & 0.720 & 0.479 \\
        GLD (Ours) &  \ding{51} & 14.542 & 0.288 & 0.504 & 0.589 & 15.97 & 1.071 & 0.630 & 0.406 \\
        \bottomrule
        \end{tabular}%
    }
    \vspace{-10pt}
\end{table*}

\subsection{Additional Qualitative Results}
\label{app:additional_results}

\begin{figure}[!h]
    \centering
    \vspace{-15pt}

    \includegraphics[width=\textwidth]{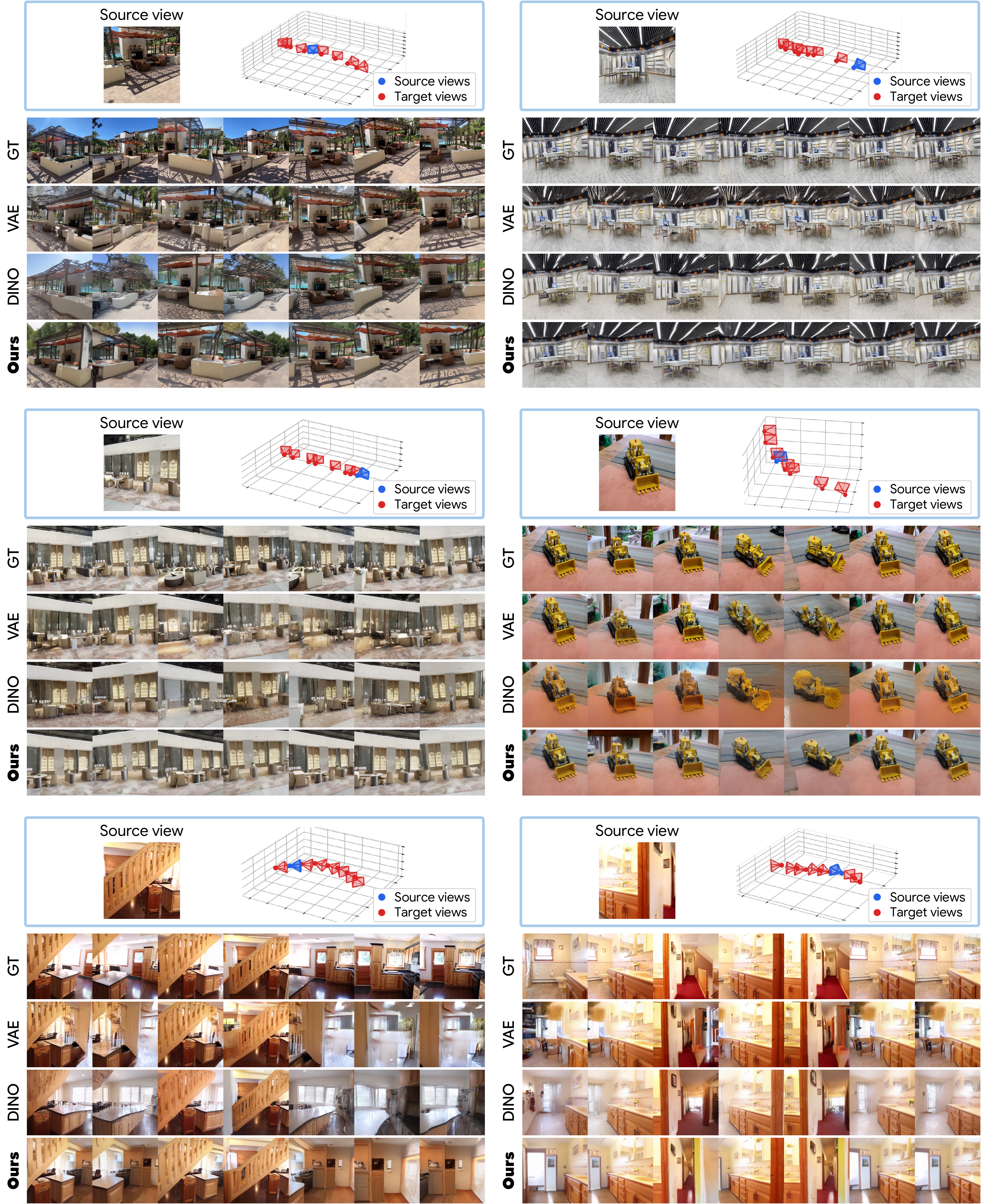}
    \caption{\textbf{Additional qualitative results from \textit{one} source views.}}
    \label{supfig:suppl_qual_1v}
    \vspace{-30pt}

\end{figure}

\begin{figure}[!h]
    \centering
    \includegraphics[width=\textwidth]{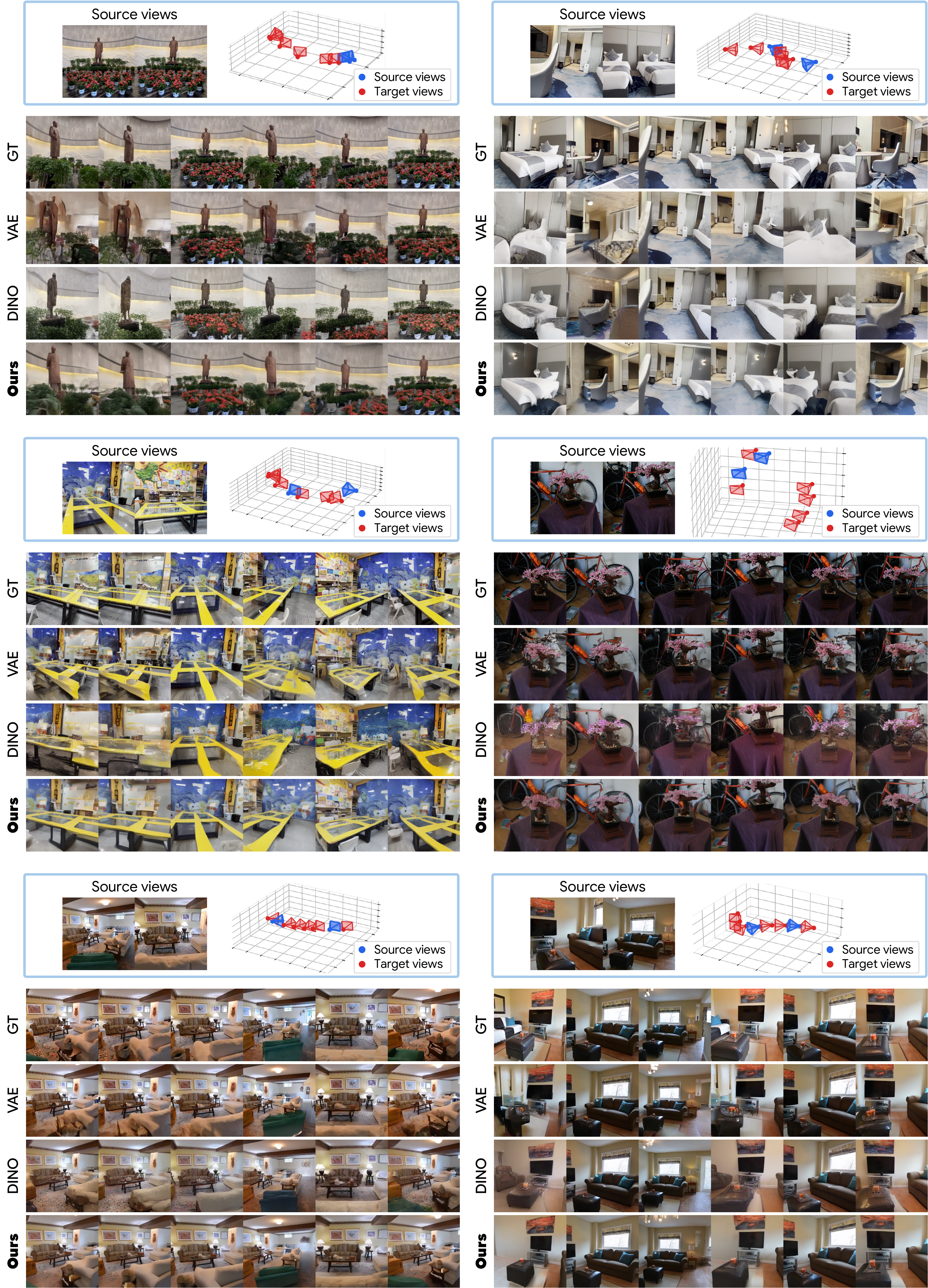}
    \caption{\textbf{Additional qualitative results from \textit{two} source views.}}
    \label{supfig:suppl_qual_2v}
\end{figure}

\begin{figure}[!h]
    \centering
    \includegraphics[width=\textwidth]{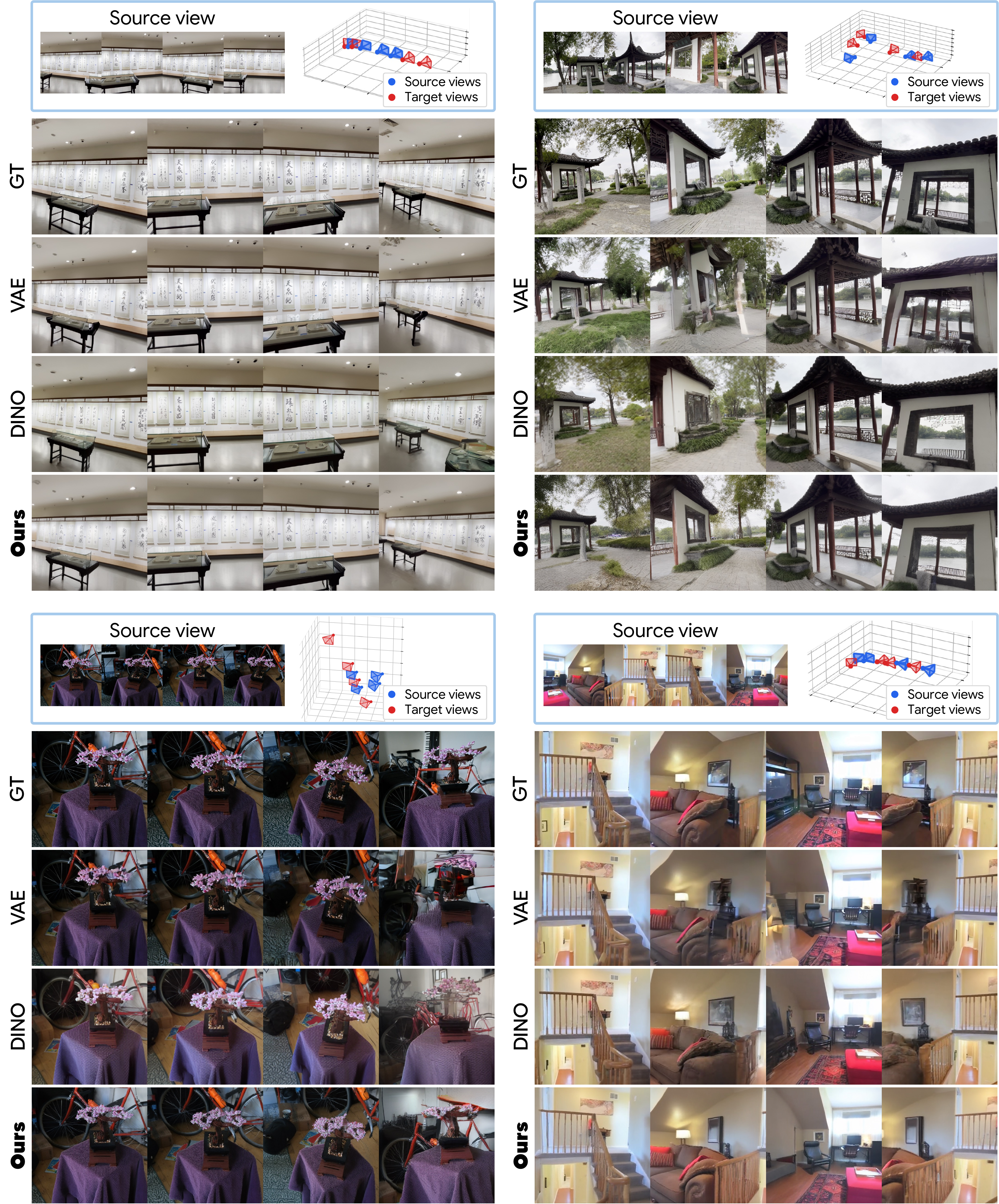}
    \caption{\textbf{Additional qualitative results from \textit{four} source views.}}
    \label{supfig:suppl_qual_4v}
\end{figure}

We present additional qualitative results for novel view synthesis using 1, 2, and 4 source views in \cref{supfig:suppl_qual_1v}, \cref{supfig:suppl_qual_2v}, and \cref{supfig:suppl_qual_4v}, respectively. We compare our method with the VAE and DINO baselines across all settings. As shown in the figures, our method consistently produces multi-view results that are more 3D consistent and photorealistic than those of the baselines, while better preserving geometric structure and appearance across viewpoints.

\clearpage

\clearpage

\subsection{Additional 3D Visualizations}
\label{app:3d_vis}
\begin{figure}[h]
    \centering
    \includegraphics[width=\textwidth]{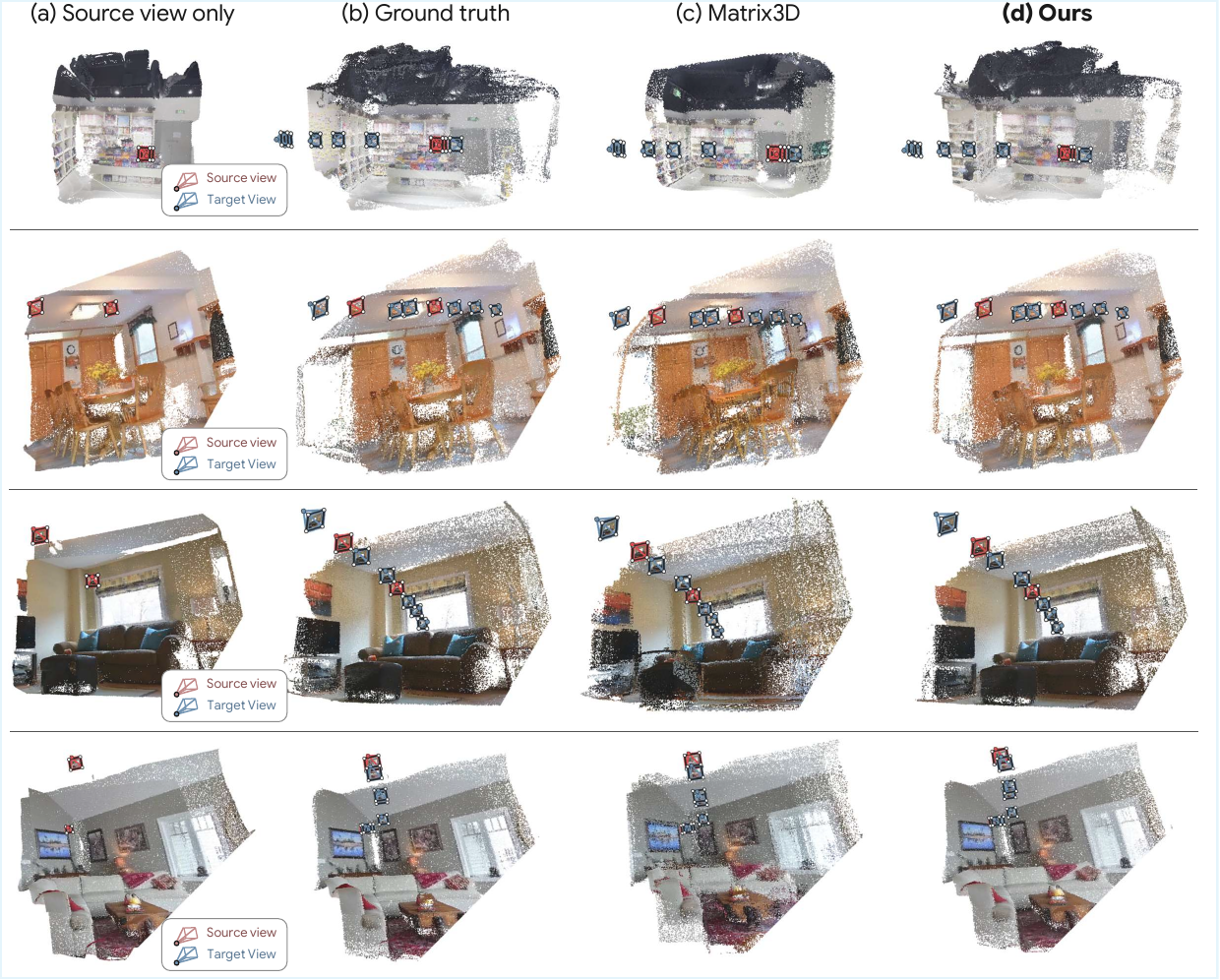}
    \caption{\textbf{Additional 3D Visualizations.} 3D reconstruction comparisons between Matrix3D and our method. \textbf{(b) Ground truth} is reconstructed using DA3 on GT RGB images. Our method yields significantly more consistent and geometrically accurate results.}
    \label{supfig:3d_additional}
    \vspace{-10pt}

\end{figure}
In \cref{supfig:3d_additional}, we present additional 3D visualizations comparing our method against Matrix3D. Due to the absence of ground-truth depth data, the reference reconstructions, denoted as ground truth, are generated by processing the ground-truth RGB images with DA3. As illustrated in these results, our approach can produce consistent and geometrically accurate 3D reconstructions.

Specifically, in extrapolative settings where the model synthesizes unobserved regions (rows 1--3), Matrix3D often exhibits misalignment between the newly generated content and the existing source views. For instance, in the dining room scene (row 2), inadequate cross-view consistency in Matrix3D results in a duplicated clock on the wall and blurry chairs. Similarly, in the third row, the generated unseen regions, such as the TV and picture frames on the left wall, do not align properly with the previously observed regions, leading to structural discontinuities. In contrast, our method generates these unobserved areas while maintaining robust 3D consistency. Furthermore, in the interpolative setting (row 4), our method synthesizes precise viewpoints and RGB-D maps that accurately correspond with the source views, preserving rigid structures and object boundaries without noticeable misalignment artifacts.

% =================================================================
%  D.  Discussion
% =================================================================
\section{Additional Discussion}
\label{app:discussion}

\subsection{Analysis of Geometric Correspondences in Diffusion Features}

CAMEO~\cite{kwon2025cameo} shows that cross-view correspondence in internal attention maps is a strong indicator of multi-view generation quality, and that models exhibiting stronger correspondence on their 3D attention maps tend to produce more geometrically consistent outputs. To examine whether this relationship holds across different latent spaces, we measure cross-view correspondence of the 3D attention maps at each layer for diffusion models trained in the DA3, VAE, and DINO latent spaces.

Specifically, given an image pair $(\mathcal{I}_1, \mathcal{I}_2)$, we extract the query $Q_1^l$ and key $K_2^l$ from the 3D attention module at each layer $l$, following CAMEO~\cite{kwon2025cameo}. Correspondences are estimated by matching each query descriptor in $Q_1^l$ to its nearest neighbor in $K_2^l$ using cosine distance. We evaluate the resulting correspondences on ScanNet~\cite{dai2017scannet} using PCK, following the protocol of Probe3D~\cite{el2024probing}. We report results for $\mathcal{M}_1$ trained in the DA3 latent space. The results are shown in \cref{fig:suppl_corr}.

\input{fig/suppl/suppl_correspondence}

Among the three latent spaces, the model trained in the DA3 latent space exhibits the strongest cross-view correspondence across nearly all layers, with the largest margin appearing in the decoder blocks. This is consistent with CAMEO's finding~\cite{kwon2025cameo} that stronger internal correspondence correlates with higher-quality multi-view generation, corroborating the superior geometric consistency observed in the final outputs (Tab.~3). Furthermore, this suggests that the DA3 latent space provides a representation that facilitates the diffusion model's learning of cross-view correspondence, rather than merely enabling better decoding after generation.

We also observe a clear asymmetry between the conditional encoder and velocity decoder. Across all three latent spaces, correspondence is essentially absent throughout the condition encoder and emerges only in the velocity decoder, where it increases sharply and peaks at intermediate blocks (layers 31--32). This suggests that the encoder primarily preserves per-view conditioning information, while cross-view geometric correspondence is established in the decoder through 3D attention.

% Known failure modes: severe occlusion, areas with very sparse coverage,
% extreme lighting changes, large temporal gap between reference and target,
% hallucinated content in regions without reference coverage.

\subsection{Computational Cost}
\input{table/sup_comp}
We report the sampling time of each model in \cref{tab:latency}(A). The model operating in the VAE latent space achieves the fastest generation speed due to its smaller token count, whereas GLD is the slowest because it requires two sampling stages to obtain the complete feature set. We further provide a latency breakdown of each module in GLD in \cref{tab:latency}(B). The results show that obtaining level 2 and level 3 features via propagation is more efficient than generating them independently, supporting our design choice to avoid explicit generation of deeper feature levels.

\subsection{Limitations}

\label{app:limitations}
\begin{figure}[!t]
    \centering
    \includegraphics[width=\textwidth]{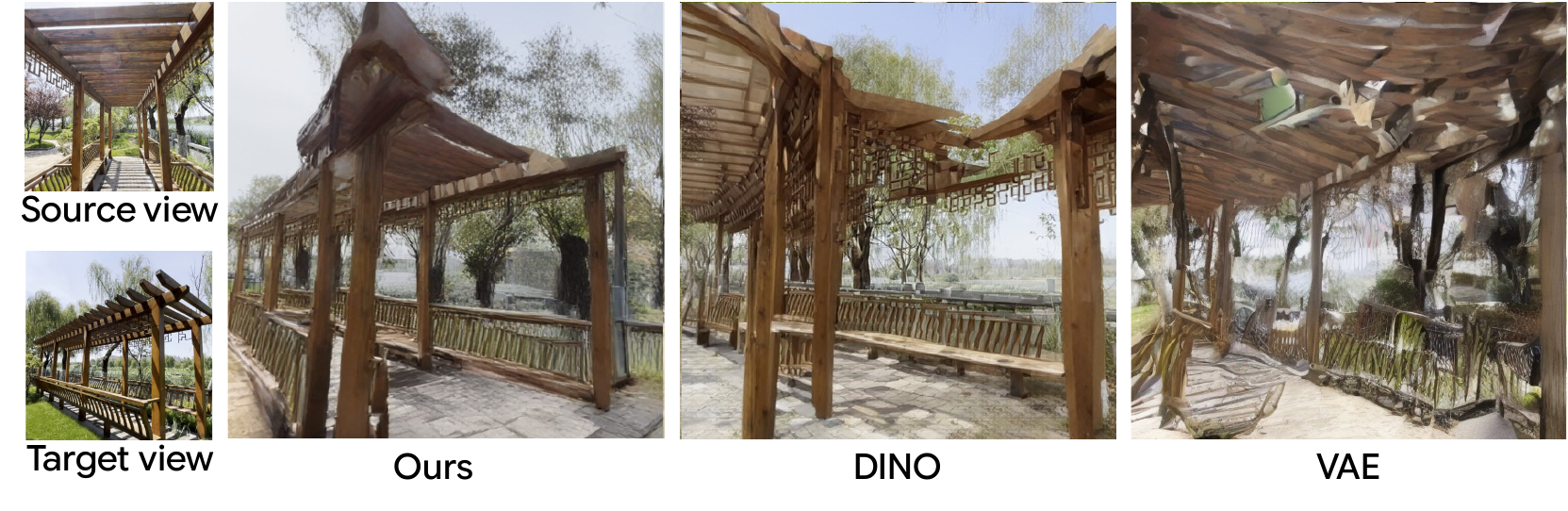}
    \vspace{-20pt}
    \caption{\textbf{Failure cases.}}
    \label{supfig:failure}
    \vspace{-10pt}
\end{figure}

While Geometric Latent Diffusion (GLD) demonstrates robust multi-view consistency, a few challenging scenarios remain, as illustrated in \cref{supfig:failure}. Specifically, in cases of severe occlusion or very sparse spatial coverage, the model may hallucinate content or produce artifacts in regions entirely unobserved by the reference views. Additionally, extreme lighting changes or large temporal gaps between the reference and target inputs can make it difficult to establish reliable cross-view correspondences.

%% file: fig/suppl/full_architecture.tex
\begin{figure}[h]
  \centering
  \includegraphics[width=1.0\linewidth]{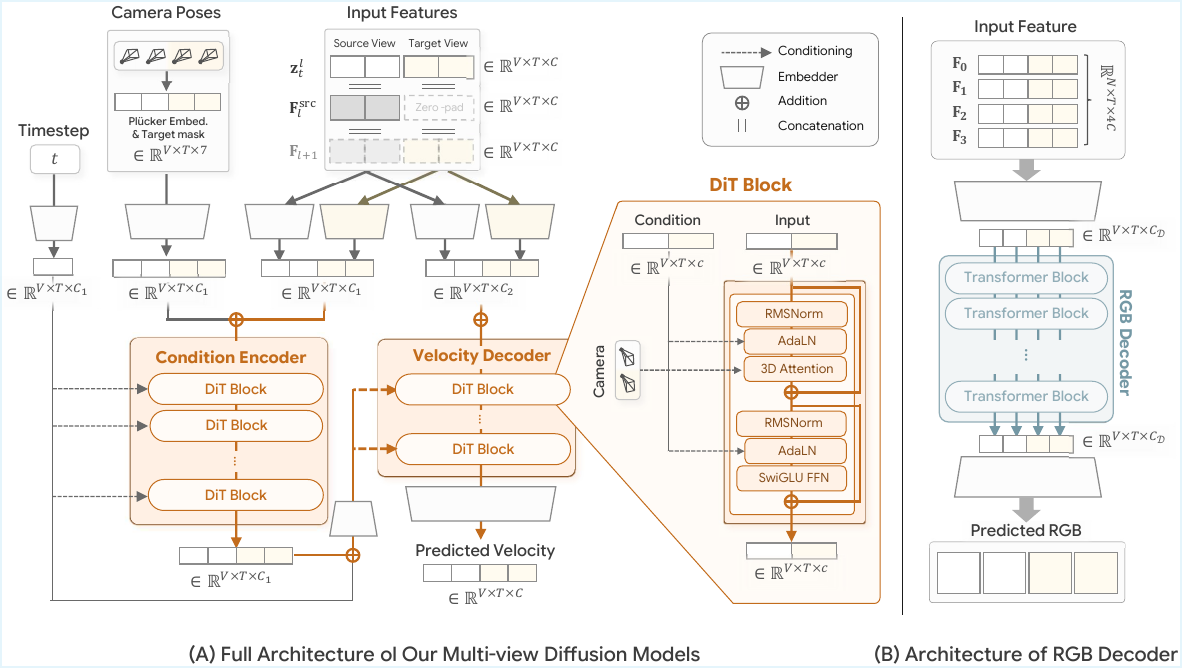}
  \vspace{-15pt}
  \caption{\textbf{Architecture details.} (A) Our multi-view diffusion model, featuring a condition encoder and a velocity decoder conditioned on camera poses and multi-view features. (B) The RGB decoder architecture that maps the multi-level latent features back to the pixel space.}
  \label{fig:suppl_full-architecture}
  % \vspace{-12pt}
\end{figure}

%% file: fig/suppl/suppl_correspondence.tex
\begin{figure}[h]
  \centering
  \includegraphics[width=0.7\linewidth]{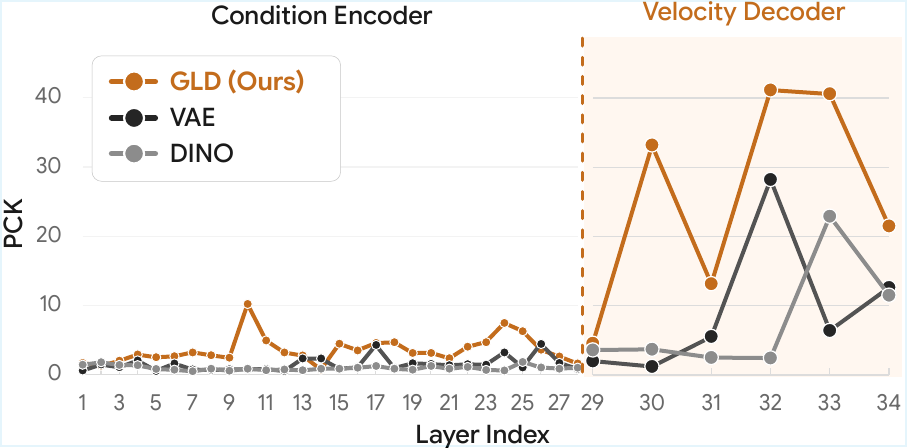}
  \vspace{-7pt}
  \caption{\textbf{Geometric correspondence of trained multi-view diffusion models.}}
  \label{fig:suppl_corr}
  \vspace{-12pt}
\end{figure}

%% file: table/sup_comp.tex
\begin{wraptable}[14]{r}{0.48\textwidth}
  \centering
  \vspace{-50pt}
  \caption{\textbf{Inference latency.} Measured on a single RealEstate10K24~\cite{re10k} scene.}
  \label{tab:latency}
  \setlength{\tabcolsep}{4pt}
  \footnotesize
  \renewcommand{\arraystretch}{0.95}

  \text{(A) Overall comparison}\\[2pt]
  \resizebox{\linewidth}{!}{%
  \begin{tabular}{@{}lrr@{}}
    \toprule
    Method & Sampling (s) & Decode (s) \\
    \midrule
    DINO       & 35.2 & 0.41 \\
    VAE        & 28.0 & 0.50 \\
    GLD (ours) & 66.1 & 0.43 \\
    \bottomrule
  \end{tabular}
  }

  \vspace{4pt}

  \text{(B) GLD runtime breakdown}\\[2pt]
  \resizebox{0.8\linewidth}{!}{%
  \begin{tabular}{@{}lr@{}}
    \toprule
    Phase & Time (s) \\
    \midrule
    Lv.1 sampling              & 37.8 \\
    Lv.1 $\rightarrow$ Lv.2, Lv.3 prop. & 0.15 \\
    Lv.1 $\rightarrow$ Lv.0 sampling  & 28.4 \\
    RGB decoding             & 0.43 \\
    \midrule
    Total                    & 66.8 \\
    \bottomrule
  \end{tabular}
  }

\end{wraptable}